%% file: acl2023.tex
\pdfoutput=1
\documentclass[11pt]{article}

\usepackage[]%[review]
{ACL2023}

\usepackage{times}
\usepackage{latexsym}
\usepackage{enumitem}
\usepackage{float}
\usepackage{acro}
\usepackage{dsfont}
\usepackage{relsize}
\usepackage{hyperref}
\usepackage{enumitem}

\usepackage{pgfplots}
\pgfplotsset{compat=1.18}

\usepackage{soul}
\usepackage{xcolor}

\newcommand{\ctext}[3][RGB]{%
  \begingroup
  \definecolor{hlcolor}{#1}{#2}\sethlcolor{hlcolor}
  \hl{#3}
  \endgroup
}

\newcommand{\gls}[1]{\texttt{#1}}

\usepackage[T1]{fontenc}
\usepackage[normalem]{ulem}

\usepackage[utf8]{inputenc}

\usepackage{microtype}
\usepackage{subcaption}
\usepackage{tabularx}
\usepackage{tabularray}
\usepackage{listings}
\usepackage{inconsolata}
\usepackage{graphicx}
\usepackage[size=tiny]{todonotes}
\newcommand{\yc}[1]{\textcolor{black}{#1}}
\newcommand{\ycr}[1]{\textcolor{black}{#1}}
\newcommand{\se}[1]{\textcolor{black}{#1}}
\newcommand{\sen}[1]{\textcolor{black}{#1}}
\newcommand{\ma}[1]{\textcolor{black}{#1}}
\newcommand{\maa}[1]{\textcolor{black}{#1}}
\newcommand{\manew}[1]{\textcolor{black}{#1}}

\newcommand{\ArgSum}{ArgSum}

\newcommand{\ArgQ}{ArgQ}
\newcommand{\ArgKPto}{ArgKP21}
\newcommand{\BarH}{BarH}
\newcommand{\SMatchToPr}{SMatchToPr}
\newcommand{\MCArgSum}{MCArgSum}
\newcommand{\USKPM}{USKPM}

\usepackage{newunicodechar}
\usepackage{amssymb}
\usepackage{amsmath}
\usepackage{booktabs}
\usepackage{array}

\newcommand\Warning{%
 \makebox[1.4em][c]{%
 \makebox[0pt][c]{\raisebox{.1em}{\small!}}%
 \makebox[0pt][c]{\color{red}\Large$\bigtriangleup$}}}%

\newunicodechar{⚠}{\Warning}

\DeclareAcronym{argsum}{short = ArgSum, long = Argument Summarization}
\DeclareAcronym{am}{short = AM, long = Argument Mining}
\DeclareAcronym{llms}{short = LLMs, long = Large Language Models}

\title{Argument Summarization and \ma{its} Evaluation in the \\ Era of Large Language Models}

\author{Moritz Altemeyer$^{a}$, Steffen Eger$^{b,d}$, Johannes Daxenberger$^{c}$, \\ \textbf{Yanran Chen$^{b,d}$, Tim Altendorf, Philipp Cimiano$^{a}$, Benjamin Schiller$^{c}$} \\
$^a$Bielefeld University,
$^b$Natural Language Learning \& Generation (NLLG), $^c$summetix GmbH, \\ 
 $^d$University of Technology Nuremberg\\
}

\begin{document}
\maketitle

\begin{abstract}
Large Language Models (LLMs) have revolutionized various Natural Language Generation (NLG) tasks, including Argument Summarization (ArgSum), a key subfield of Argument Mining. 
\manew{This paper investigates the integration of state-of-the-art LLMs into ArgSum systems and their evaluation.} In particular, we propose a novel prompt-based evaluation scheme, and validate it through a novel human benchmark dataset. Our work makes three 
\maa{main} contributions: \maa{(i)} the integration of LLMs into existing ArgSum \manew{systems,} 
\maa{(ii)} the development of 
\manew{two} new \maa{LLM-based} ArgSum \manew{systems,}
benchmarked against prior methods, and \maa{(iii)} the introduction of an advanced LLM-based evaluation scheme. We demonstrate that the use of LLMs substantially improves both the generation and evaluation of argument summaries, achieving state-of-the-art results and advancing the field of ArgSum. \yc{We also show that among the four LLMs integrated in (i) and (ii), Qwen-3-32B, despite having the fewest parameters, performs best, even surpassing GPT-4o.}
\end{abstract}

\input{sections/1_introduction_may}
\input{sections/2_related_work}
\input{sections/3_setup}
\input{sections/4_results_may}

\input{sections/6_conclusions_may}
\input{sections/7_ack_limit_ethic}

\section*{Acknowledgements}

We thank the anonymous reviewers for their thoughtful feedback, which helped improve the paper. 
The NLLG Lab gratefully acknowledges support from the Federal Ministry of Education and Research (BMBF) via the research grant
“Metrics4NLG” and the German Research Foundation (DFG) via the Heisenberg Grant EG 375/5-1. This project has been conducted as part of the EMCONA (The Interplay of Emotions and Convincingness in Arguments) project, which is funded
by the DFG (project EG-375/9-1). 

\bibliography{anthology,custom,moritz}
\bibliographystyle{acl_natbib}

\input{sections/100_appendix}

\end{document}

%% file: sections/1_introduction_may.tex
\section{Introduction} \label{sec:introduction}
\ac{llms} have 
significantly transformed various Natural Language 
Processing (NLP) and Generation (NLG) 
problems. 
Their remarkable capabilities in understanding and generating human-like text  
promise new avenues for challenging tasks such as \textit{\ac{argsum}}, a subfield of Argument Mining that focuses on distilling the essence of multiple arguments into concise representations \cite{friedman-etal-2021-overview}.\footnote{\maa{While past work on summarizing argumentative texts conveys different understandings of the task at hand, our understanding aligns with Key Point Analysis, introduced by \citet{bar-haim-etal-2020-arguments,bar-haim-etal-2020-quantitative}.}}

With only a few recent exceptions \cite{li-etal-2024-side,10.1007/978-3-031-63536-6_20}, however, 
\ac{argsum} has up-to-date been mostly tackled with pre-LLM solutions, such as clustering techniques \ma{and} 
earlier-generation pre-trained language models 
{\cite{misra-etal-2016-measuring,reimers-etal-2019-classification,ajjour-etal-2019-modeling,wang-ling-2016-neural,schiller-etal-2021-aspect,bar-haim-etal-2020-arguments,bar-haim-etal-2020-quantitative,alshomary-etal-2021-key,li-etal-2023-hear}.

Thus, there is an urgent need for systematic analysis to understand how 
LLMs 
can be effectively utilized for  
the generation and evaluation of argument summaries. \ma{This includes integrating LLMs into ArgSum frameworks to comprehensively assess their performance and developing suitable prompt-based evaluation schemes.}

In this work, we aim to fill 
this 
gap by extensively exploring how LLMs can be 
leveraged for the 
ArgSum process, both for generating argument summaries and for their evaluation. 
Our 
contributions are:
(i) We integrate LLMs into existing \ArgSum{} systems, \yc{showing substantial performance gains.} 
(ii) We introduce \manew{two} 
new \maa{LLM-based} \ArgSum{} \manew{systems,}
\yc{
showing \manew{their} 
superiority over existing approaches.}
\yc{(iii) We show that among the four LLMs used in (i) and (ii), the smallest one, Qwen3-32b, performs best, even surpassing GPT-4o, while LLaMA-3.3-70B consistently underperforms.} 
(iv) We provide a new ArgSum evaluation dataset with 
human evaluation scores. 
(v) We develop a prompt-based \ArgSum{} evaluation scheme,  \yc{showing stronger correlation with human judgments than existing automatic evaluation metrics.}\footnote{Code and Data available at \url{https://github.com/NL2G/argsum}} 

%% file: sections/2_related_work.tex
\section{Related Work} \label{sec:related}

\subsection{Argument Summarization}
Automatic Text Summarization (ATS) aims to condense \maa{the} key ideas from one or more documents into a concise summary \cite{10.1162/089120102762671927}, while minimizing redundancy \cite{7944061}. 
While abstractive summarization generates a summary \ma{including}
%with 
text \ma{units} that \ma{do} not \ma{necessarily appear}
in the source text, extractive summarization identifies the most important parts of a document and assembles them into a summary \cite{app13137620}. 
\ma{ATS consists of}
several sub-areas like News Summarization \cite{8336568}, Legal Document Summarization \cite{ANAND20222141}, Scientific Paper Summarization \cite{ZHANG201888}, and \ac{argsum} \citep{bar-haim-etal-2020-arguments,bar-haim-etal-2020-quantitative}. 
\se{Our focus is the latter.}

\maa{\citet{misra-etal-2016-measuring}, \citet{reimers-etal-2019-classification} and \citet{ajjour-etal-2019-modeling} treat the task of summarizing arguments as a clustering problem without providing easy-to-understand textual summaries.} \citet{wang-ling-2016-neural} \ma{frame ArgSum as}
claim generation, where a collection of argumentative sentences is summarized by generating a one-sentence abstractive summary that addresses the shared opinion of the inputs. \citet{schiller-etal-2021-aspect} 
\ma{present an aspect-controlled argument generation model that enables an abstractive summatization of arguments.}

\begin{figure}[t!]
    \centering
    \includegraphics[width=1\columnwidth]{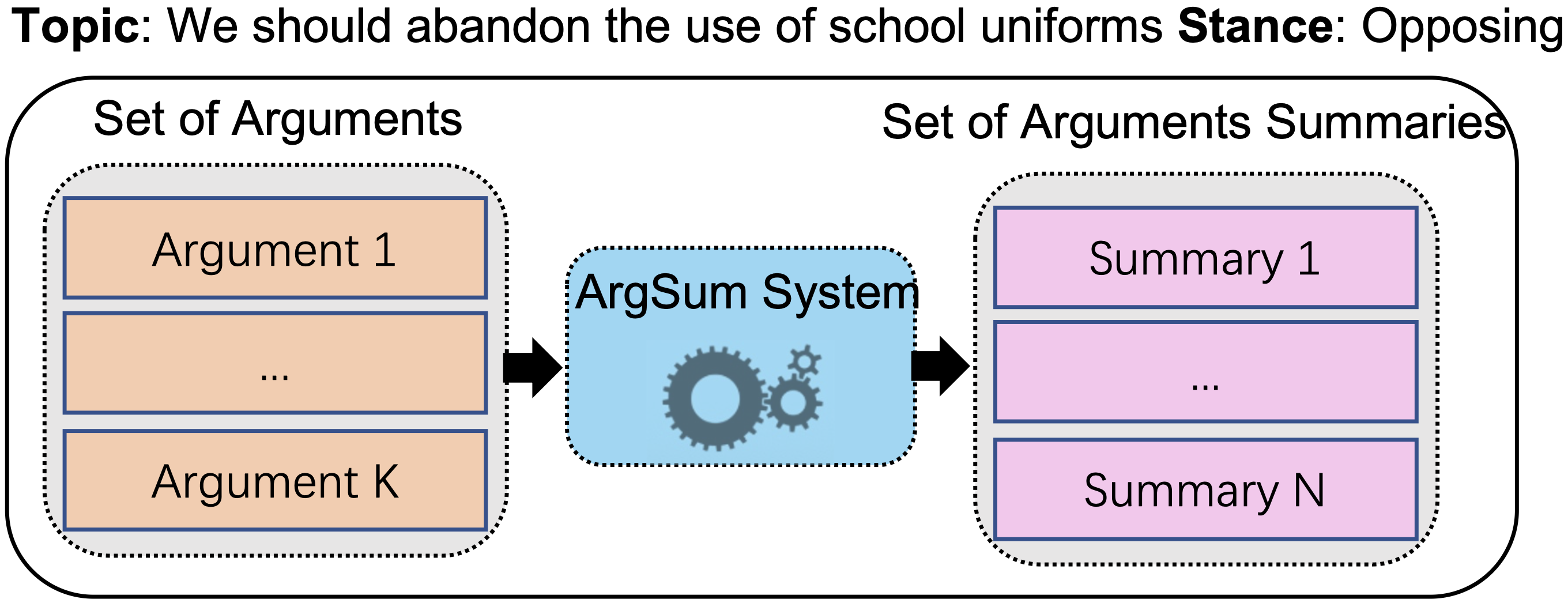}
    \caption{General procedure of ArgSum, where a set of $K$ arguments on a certain debate topic and stance (example taken from \citet{friedman-etal-2021-overview}) is transformed to a set of $N$ argument summaries.  
    It is expected that $K \gg N$ applies.}
    \label{fig:argsumexample}
    \vspace{-.5cm}
\end{figure}

\maa{Our understanding of ArgSum is in line with Key Point Analysis \ma{(KPA)}, introduced by \citet{bar-haim-etal-2020-arguments,bar-haim-etal-2020-quantitative}, and is displayed in Figure~\ref{fig:argsumexample}.} 
\manew{They aim to create a summary consisting of the most important key points on a specific debate topic and stance by extracting them from a potentially large collection of relevant arguments.}
\maa{Then, each source argument is classified according to the most suitable key point.} \maa{\citet{alshomary-etal-2021-key}} perform the key point extraction by utilizing a variant of PageRank \citep{Page1998PageRank}. 
\citet{li-etal-2023-hear} \ma{extend KPA with a clustering-based and abstractive approach}\maa{, using grouped arguments as input for a generation model to create key points.}
\citet{khosravani-etal-2024-enhancing} \maa{introduce a clustering-based and extractive approach, selecting the most representative argument within each cluster as a key point, determined by a supervised scoring model.} 

\subsection{Evaluating NLG Systems}

\ma{While} 
automatic \ma{evaluation} metrics such as BLEU \cite{10.3115/1073083.1073135} and ROUGE \cite{lin-2004-rouge} 
correlate poorly with human judgments \cite{novikova-etal-2017-need}, 
\ma{pre-trained} transformer-based language models provide a more nuanced 
\ma{assessment} of the performance of NLG systems \cite{celikyilmaz2021}.
BERTScore \cite{bertscore} and MoverScore \cite{zhao-etal-2019-moverscore} are reference-based metrics that leverage pre-trained embeddings obtained from BERT-based models. While BERTScore is based on the computation of cosine-similarities between \ma{the hypothesis and the reference,}
MoverScore determines an evaluation score by computing the Word Mover’s Distance \cite{pmlr-v37-kusnerb15} \ma{between both.}
BARTScore \cite{bartscore} is based on the pre-trained sequence-to-sequence model BART and treats the evaluation task as a problem of text generation. \maa{BLEURT \cite{sellam-etal-2020-bleurt} is reference-based and consists of a BERT-based model that is fine-tuned to predict human ratings in such a way that the metric is robust to quality shifts.} MENLI \cite{chen-eger-2023-menli} frames the evaluation task as a problem of Natural Language Inference (NLI), showing improved robustness.  
The most recent approaches \se{to evaluation} are LLM-based metrics, \ma{which can be leveraged in various ways:} 
by comparing embeddings in terms of their cosine similarity \cite{es-etal-2024-ragas}, by determining the sequence probability of the hypothesis given the respective source/reference \cite{fu-etal-2024-gptscore}, by utilizing suitable prompting strategies \cite{kocmi-federmann-2023-large,liu-etal-2023-g,fernandes-etal-2023-devil,leiter-eger-2024-prexme,larionov-eger-2025-promptoptme}, or by applying task-specific fine-tuning \cite{pandalm2024,xu-etal-2023-instructscore,zhu2023judgelm}. Some works show promising zero-shot results that are on-par with 
human-judgement 
\cite{leiter-etal-2023-eval4nlp,10.1145/3641289}.

In this work, we 
leverage LLMs to evaluate ArgSum systems, which is different from evaluation of classical text generation systems, 
requiring different dimensions of evaluation (e.g., coverage and redundancy) and different mechanisms (e.g., ArgSum requires to compare \maa{$m$ reference summaries to $n$ generated summaries}). 
To this end, we apply an LLM-based prompting approach 
\maa{and compare it with two existing ArgSum evaluation frameworks.}

\subsection{Argument Summarization and Evaluation with LLMs}

Among the works that use LLMs for argument summarization or evaluation, \citet{10.1007/978-3-031-63536-6_20} use snippet generation and neutralization (a mix of extractive summarization
and LLM prompting with reinforcement learning) for ArgSum in the context of argument search. They evaluate their approach automatically and manually, but do not apply LLMs for the evaluation. While their ArgSum task is different from ours, they also did not assess state-of-the-art LLMs like GPT-4o.
\citet{li-etal-2024-side} apply LLMs to argumentative essay summarization. 
They test a variety of state-of-the-art LLMs to generate reference summaries which are evaluated by humans. 
Their own summarization system, however, relies on smaller, instruction fine-tuned models rather than state-of-the-art LLMs.
Different to our work, for the ArgSum evaluation, they only apply standard metrics like ROUGE.

Our work is the first \emph{systematic} study on strategies for integrating LLMs 
\manew{into} existing approaches for ArgSum and  
\manew{their} evaluation. 

%% file: sections/3_setup.tex
\vspace{-.2cm}
\section{Experimental Setup} \label{sec:setup}
\vspace{-.2cm}
\subsection{Terminology}
\vspace{-.2cm}

Most previous work on ArgSum can be categorized as either \emph{classification-based} or \emph{clustering-based} systems. Classification-based systems first generate a set of argument summaries based on all available arguments. In a second step, they match each source argument to the most appropriate summary. On the other hand, clustering-based systems group all source arguments according to their similarity. Then, they generate a summary of the arguments for each cluster. In this work, we augment ArgSum systems of both types with LLMs. 
Details of those systems and how we integrate LLMs are specified in 
\S\ref{sec:ExtractiveSystems}~and~\S\ref{sec:Clustering-based_Systems}.

The systems we assess 
use two types of \ma{tools to perform ArgSum.} 
\ma{While \emph{Quality Scorers} assess the quality of an argument, \emph{Match Scorers} determine how well an argument and a summary match.}
\ma{Both} are realized by transformer-based language models that take task-specific textual inputs and output a respective score. \ma{The ArgSum systems considered in this work utilize Quality Scorers that are fine-tuned on the IBM-ArgQ-Rank-30kArgs (\ArgQ) dataset by \citet{DBLP:conf/aaai/GretzFCTLAS20}. The corresponding Match Scorers are fine-tuned on the ArgKP-2021 (\ArgKPto) dataset by \citet{friedman-etal-2021-overview}.} \maa{We provide details on the required model fine-tuning for the ArgSum systems discussed in %Sections~
\S\ref{sec:ExtractiveSystems}~and~\S\ref{sec:Clustering-based_Systems} %are %collected 
in Appendix \ref{sec:Details on Model Fine-tuning}.}

\subsection{\ma{Classification-based} Systems} 
\label{sec:ExtractiveSystems}
\begin{figure*}[t!]
    \centering
    \includegraphics[width=\textwidth]{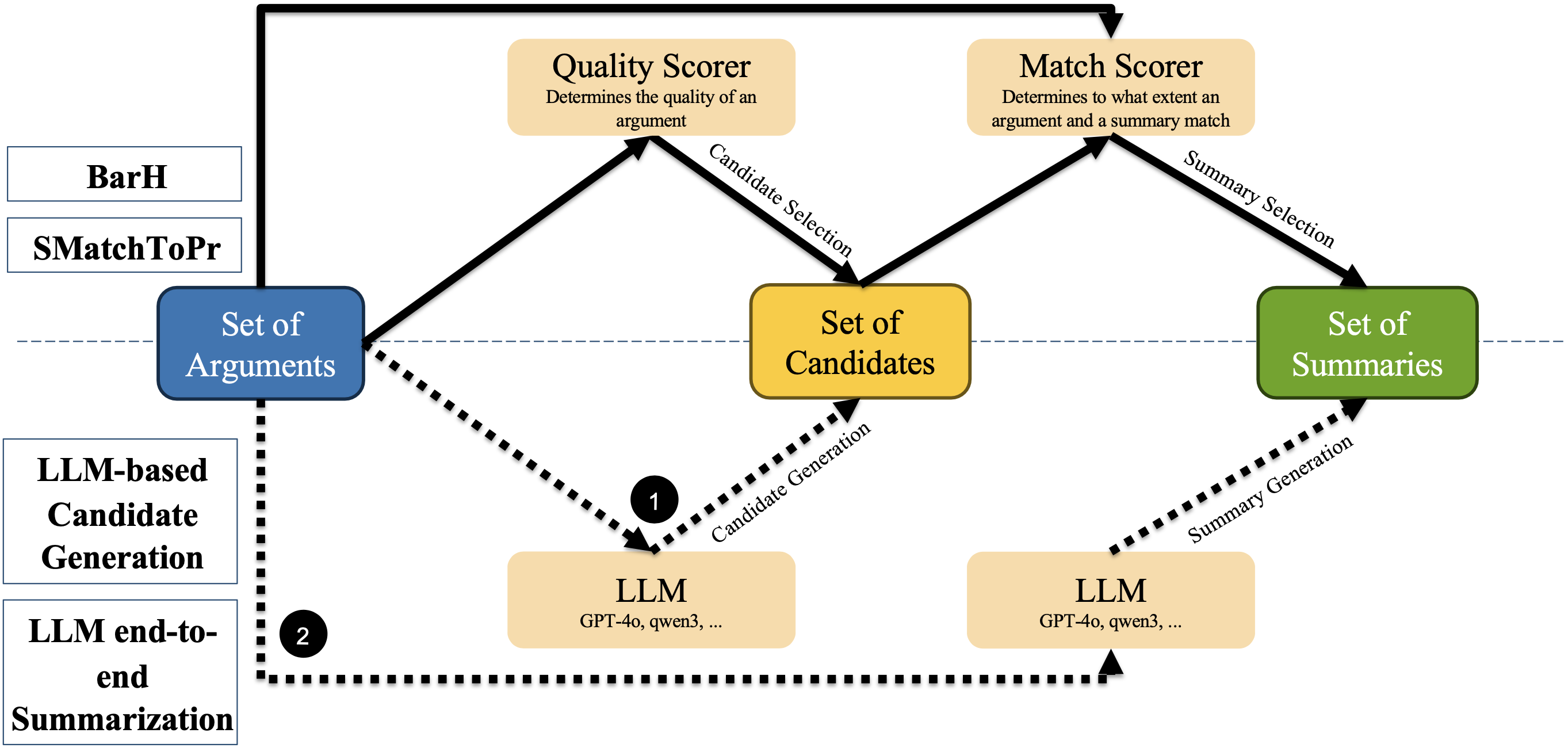}
    \vspace{-.5cm}
    \caption{Overview Classification-based Systems and LLM end-to-end Summarization. Dashed lines indicate (1) LLM integration into classification-based ArgSum systems and (2) LLM end-to-end generation of argument summaries.}
    \label{fig:overview-clf-based}
    \vspace{-.5cm}
\end{figure*}

We consider two classification-based 
\ma{ArgSum} systems, which 
performed best in the Key Point Generation Track at the 2021 Key Point Analysis Shared Task \citep{friedman-etal-2021-overview}. Both approaches and the integration of LLMs into them are visualized in Figure~\ref{fig:overview-clf-based}. 

\paragraph{\BarH}
To determine a set of potential argument summaries, referred to as candidates, BarH \citep{bar-haim-etal-2020-quantitative} exclude from consideration those arguments that consist of multiple sentences, exceed a token threshold $n$, or start with pronouns. The remaining arguments are then scored by a Quality Scorer, and only those with a score above a threshold $t_q$ are selected as candidates. Subsequently, BarH applies a Match Scorer to match the excluded source arguments to the best fitting candidates. After ranking the candidates according to their number of matches, BarH minimizes redundancy by removing candidates whose match score with a higher-ranked candidate exceeds a threshold $t_m$. The remaining candidates are understood as the final argument summaries.

\paragraph{\SMatchToPr}
To identify argument summary candidates, SMatchToPr \citep{alshomary-etal-2021-key} split multi-sentence arguments into individual sentences and exclude from consideration those that fall outside the token range $[n_{Min}, n_{Max}]$ or begin with pronouns. The remaining arguments that receive a quality score above a threshold $t_{q}$ are considered as candidates. To rank these candidates, a variant of PageRank \citep{Page1998PageRank} is applied, where candidates are represented as nodes in an undirected graph and match scores between candidate pairs serve as edge weights. Only nodes with edge weight above a threshold $t_n$ are connected. Based on the resulting graph, an importance score is calculated for each candidate. Then, SMatchToPr minimizes redundancy by removing candidates whose match score with a higher-ranked candidate exceeds a threshold $t_m$. This results in the final set of argument summaries. 

\paragraph{LLM-based Candidate Generation} 
Given a set of arguments on a certain debate topic and stance, we apply a zero-shot prompting approach to instruct an LLM to generate \manew{a set of candidates, which is then further processed as usual in both BarH and SMatchToPr}
\maa{(see Appendix \ref{subsec:LLM Prompting - Classification-based Systems}).}

\subsection{Clustering-based Systems}\label{sec:Clustering-based_Systems}

\begin{figure*}[t!]
    \centering
    \includegraphics[width=\textwidth]{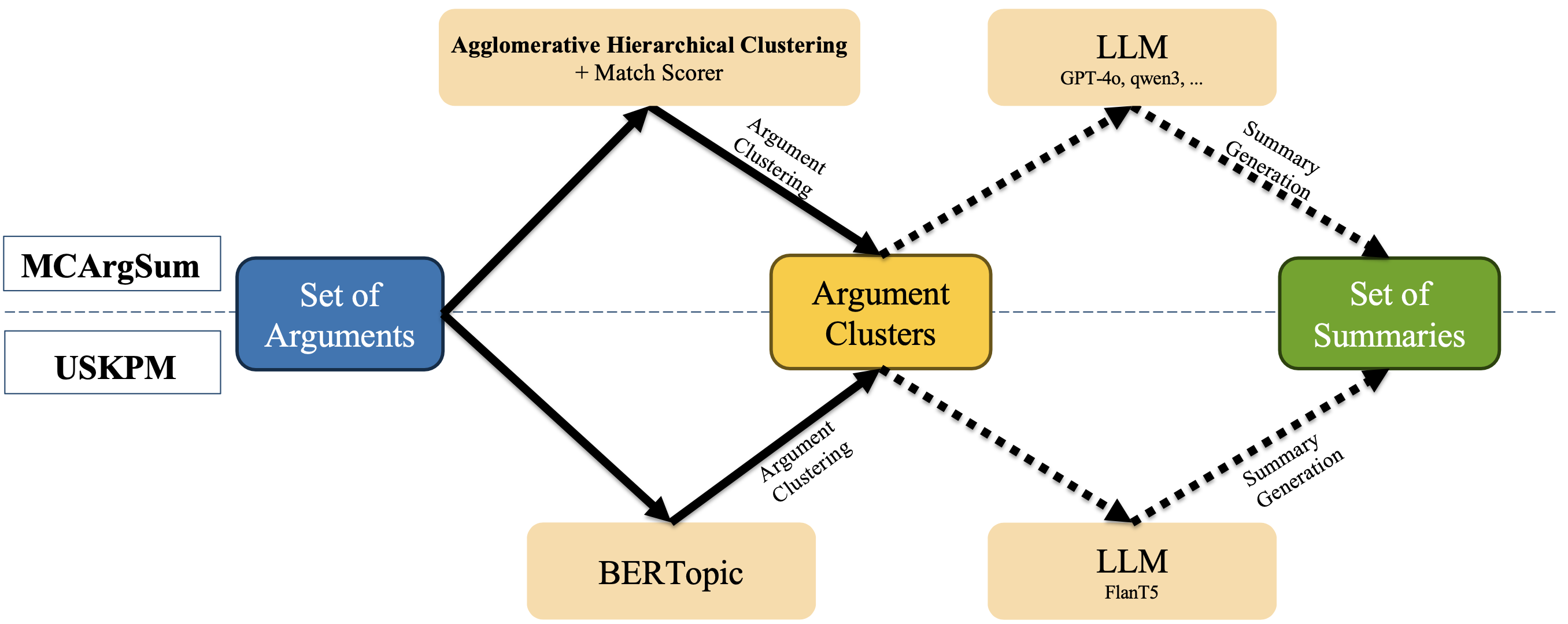}
    \vspace{-.5cm}
    \caption{Overview Clustering-based Systems. Dashed lines indicate LLM integration for \manew{cluster summarization.}
    }
    \label{fig:overview-clustering-based}
    \vspace{-.5cm}
\end{figure*}

We consider an approach from \citet{li-etal-2023-hear} which demonstrated comparable performance to BarH and SMatchToPr. Further, we propose a new ArgSum approach that utilizes a Match Scorer for argument clustering \manew{and an LLM for cluster summarization}. Both approaches are visualized in Figure~\ref{fig:overview-clustering-based}.

\paragraph{\USKPM}
For clustering arguments, USKPM \citep{li-etal-2023-hear} utilizes the BERTopic framework 
\citep{grootendorst2022bertopicneuraltopicmodeling}, which involves three steps. First, contextualized sentence embeddings of the arguments are created via SBERT 
\citep{reimers-gurevych-2019-sentence}. Second, UMAP \citep{McInnes2018umap} is applied to reduce the embeddings’ dimensionality. 
Third, the clustering of the reduced embeddings is performed by HDBSCAN 
\citep{McInnes2017hdbscan}. 
Instances included in clusters with a size smaller than $c$ are considered as unclustered. Since \citet{li-etal-2023-hear} state that it is reasonable to maximize the number of clustered arguments in order to increase the representativeness of the argument summaries to be generated, Iterative Clustering (IC) is proposed. IC is about incrementally assigning unclustered arguments to the most similar cluster in terms of cosine similarity.

\sen{Finally}, USKPM uses the instruction-tuned FLAN-T5 \citep{flant5} to summarize the argument clusters, where the model input is formatted as follows: ``summarize: \{Stance\} \{Topic\} \{List of Arguments in Cluster\}”.

\paragraph{\MCArgSum}
Our own approach, \textit{MCArgSum \se{(Match Clustering based ArgSum)}}, 
combines the use of a Match Scorer for argument clustering with an LLM-based cluster summarization. It is inspired by the redundancy reduction among candidates within BarH, where a Match Scorer is utilized to identify candidates addressing the same key point. We demonstrate that a Match Scorer can also be effectively used to group arguments addressing the same main statement. While the key idea of using a Match Scorer to group arguments %a 
\se{is} \maa{also}
proposed by \citet{khosravani-etal-2024-enhancing}, our ArgSum system additional provides an abstractive summarization of argument clusters by incorporating an LLM.

Our ArgSum system utilizes Agglomerative Hierarchical Clustering \citep{Day1984EfficientAF} with the average linkage criterion in reference to \citet{reimers-etal-2019-classification} and a Match Scorer as pairwise similarity metric. To this end, we use the SBERT \citep{reimers-gurevych-2019-sentence} model all-mpnet-base-v2, fine-tuned on ArgKP21. While the threshold $m$ determines the minimum match score required between two clusters to be merged, instances included in clusters with a size smaller than $c$ are considered as unclustered.

To generate cluster summaries, our model uses LLM prompting in a zero-shot setting. We integrate a prompting strategy that summarizes all argument clusters simultaneously (global optimization). Details are given in \maa{Appendix \ref{subsec:LLM Prompting - Clustering-based Systems}.}  
After summarization, a post-processing step automatically extracts the argument summaries in the desired format.

\subsection{LLM end-to-end Summarization}\label{sec:end-to-end}

In addition to MCArgSum, we also introduce an \textit{LLM-based end-to-end argument summarization} approach. Given a set of arguments on a certain debate topic and stance, we apply zero-shot prompting to instruct an LLM to directly generate a set of argument summaries (see Figure~\ref{fig:overview-clf-based} and Appendix \ref{subsec:LLM Prompting - Classification-based Systems}).

\subsection{Evaluation}\label{sec:eval} 
Here, we describe the approaches used to evaluate ArgSum systems. These metrics are both set-based and reference-based, meaning a set of candidate summaries is compared to a set of reference summaries. 

In accordance with previous work on generating argument summaries, we assess the two evaluation dimensions of \emph{coverage} and \emph{redundancy}. Coverage refers to the extent to which a set of argument summaries captures the central talking points of a debate. Redundancy is concerned with the extent of content overlap between the individual argument summaries \citep{bar-haim-etal-2020-quantitative, alshomary-etal-2021-key, friedman-etal-2021-overview, li-etal-2023-hear, khosravani-etal-2024-enhancing}. 

\paragraph{Soft-Score} \citet{li-etal-2023-hear} introduce 
three evaluation scores: Soft-Precision (sP), Soft-Recall (sR) and Soft-F1 (sF1).
While sP finds the most suitable reference summary for each candidate summary, sR finds the most suitable candidate summary for each reference summary. To compare references and candidates, \citet{li-etal-2023-hear} utilize a semantic similarity function. The final evaluation scores in terms of sP and sR are obtained by averaging the similarity scores of the respective best matches of references and candidates. Finally, the sF1 is the harmonic mean  
of sP and sR. Formally: 

\begin{equation}
    sP = \frac{1}{n} \cdot \sum_{a_i \in A} \max_{\beta_{j} \in B} f(a_{i}, \beta_{j})
\end{equation}

\begin{equation}
    sR = \frac{1}{m} \cdot \sum_{\beta_{j} \in B} \max_{a_i \in A} f(a_{i}, \beta_{j})
\end{equation}

where $f$ is a function evaluating the semantic similarity between two summaries; $A$ and $B$ are the sets of candidate and reference summaries, with $n$ and $m$ being their respective sizes. As similarity function, \citet{li-etal-2023-hear} suggest the use of 
BLEURT \citep{sellam-etal-2020-bleurt} and BARTScore.

\paragraph{Coverage-Score}
The Coverage-Score (CS) %, proposed by 
\citep{khosravani-etal-2024-enhancing}  
assesses the coverage of a set of candidate summaries\ma{, which is defined as the proportion of reference summaries covered by them.}
Each possible pair of candidates and references is scored by a Match Scorer and classified as matching or non-matching. The former corresponds to the case in which the respective match score is above a certain threshold $t_{M}$. Finally, the CS is derived as the proportion of references with at least one matching candidate. Formally: 
\vspace{-.3cm}

\begin{equation}
\resizebox{\columnwidth}{!}{$
    CS = \frac{1}{m} %\cdot 
    \sum_{\beta_{j} \in B} \mathlarger{\mathds{1}} \left[\sum_{a_i \in A}  \mathlarger{\mathds{1}}
 \left[\textit{match}(a_{i}, \beta_{j}) > t_{M} \right] \geq 1 \right]
 $}
\end{equation}

where $\textit{match}$ indicates the match score of two summaries; $A$ and $B$ are the sets of candidate and reference summaries, $m$ is the size of $B$ and $t$ is the matching threshold.
\citet{khosravani-etal-2024-enhancing} suggest the use of the Match Scorer inherent in BarH. A recommended threshold $t_{M}$ is not provided.

\begin{figure}[t!]
    \centering
    \includegraphics[width=\columnwidth]{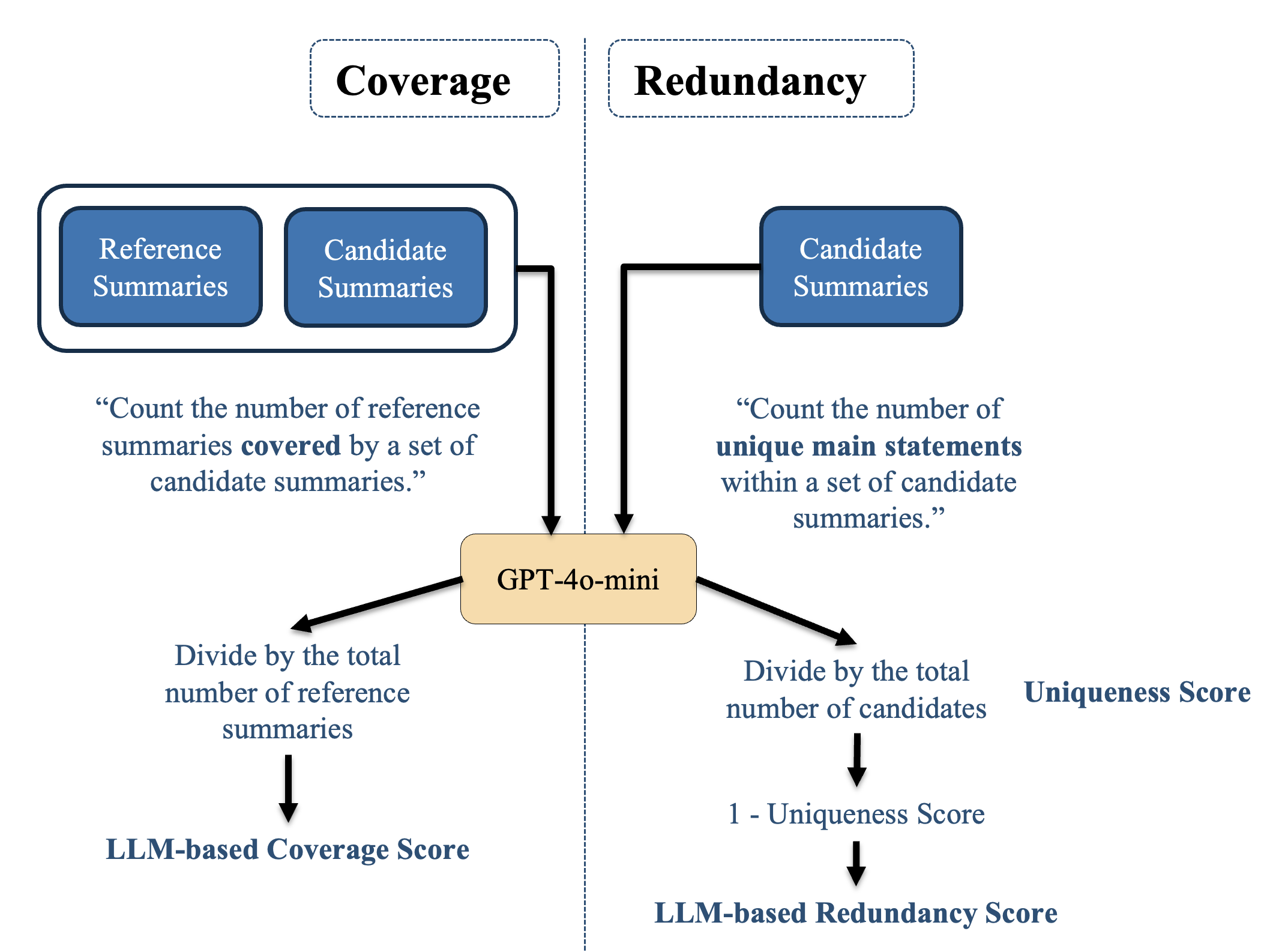}
    \caption{Overview LLM-based Evaluation. Coverage and uniqueness are assessed directly via prompting.}
    \label{fig:eval-llm-overview}
    \vspace{-.5cm}
\end{figure}

\paragraph{LLM-based}\label{subsec:LLM-based}
We introduce two %LLM 
one-shot prompting strategies for assessing ArgSum systems, focusing on the dimensions of coverage and redundancy.  
(i) We address coverage by instructing an LLM to count the number of reference summaries covered by a set of candidate summaries. Dividing this count of covered references by the total number of references results in an \textit{LLM-based Coverage Score}. (ii) To assess redundancy, we instruct an LLM to count the number of unique main statements within a set of candidate summaries. The resulting uniqueness count is limited to the total number of candidates and a uniqueness score is derived by dividing the uniqueness count by the total number of candidates. Subsequently, we derive an \textit{LLM-based Redundancy Score} as the complementary uniqueness score $(1-$Uniqueness Score$)$. 
\sen{A schematic overview is presented in Figure \ref{fig:eval-llm-overview}.}
The final LLM-based Coverage and Redundancy Scores for a certain set of candidate summaries are obtained by averaging the results of 10 evaluation runs.

\paragraph{Human Evaluation}
\label{sec:Human-Evaluation-section}
To verify the reliability of the automatic evaluation metrics, we conduct a human evaluation of 126 generated argument summaries obtained from the ArgSum systems described in \manew{\S\ref{sec:ExtractiveSystems}, \S\ref{sec:Clustering-based_Systems} and \S\ref{sec:end-to-end}, using the arguments contained in ArgKP21.}  
 We characterize a suitable set of argument summaries as consisting of succinct, non-redundant summaries that cover the main statements shared across the source arguments with adequate granularity. Thus, we assess the dimensions of coverage and redundancy, as introduced above.

The judgments are carried out by four 
\se{experienced} 
annotators with excellent knowledge within the field of NLP, especially argumentation. Initially, the four annotators are introduced to the task of ArgSum and provided with a description of the evaluation task. Guidelines can be found in Appendix \ref{sec:Human Evaluation}. The annotators are presented with a set of 
argument summaries 
and the corresponding set of reference summaries. To assess coverage, they are asked to count the number of references that are covered by the set of generated summaries. The respective coverage score is the proportion of covered references out of the total number of references. We then ask the annotators to count the number of unique main statements within the set of generated summaries (permitted count is limited to the total number of generated summaries). Based on this, we derive a uniqueness score (ranging from zero to one) as the number of unique main statements divided by the total number of generated summaries. The redundancy score is the complementary uniqueness score. 

In order to determine the inter-rater reliability, we average the Pearson correlation coefficients between each pair of the four annotators' scores for both dimensions.
We report an average correlation of 0.697 for coverage and 0.722 for redundancy, indicating that the annotations are reliable. Pairwise correlations between annotators are shown in Figure~\ref{fig:inter_rater} in the appendix.\footnote{For annotator 1 (A1), the judgments for both stances of the third topic are missing, whereas the others (A2-A4) evaluated all three topics.}

%% file: sections/4_results_may.tex
\section{Results} \label{sec:results} 
\input{results_tables/coverage}

In this section, we present the correlation of automatic metrics with human judgments in \S\ref{sec:metrics} and the evaluation of ArgSum systems in \S\ref{sec:systems}. Details on the experimental conditions, including data preprocessing, modifications to the ArgSum systems, hyperparameter settings, and hardware, are provided in Appendix \ref{Experimental conditions}.

\paragraph{Data} While ArgKP21 is used to train the Match Scorers utilized by BarH, SMatchToPr and MCArgSum, we use 
its test set to generate argument summaries in \S\ref{sec:metrics} and \S\ref{sec:systems}.  
This dataset consists of 27,519 pairs of arguments and key points, each labeled with a binary value that indicates whether the corresponding argument and key point are matching (1) or non-matching (0). While the pairs of arguments and key points cover 28 topics, each with supporting (1) and opposing (-1) stance, the dataset includes a train set of 24 topics, a development set of 4 topics and a test set of 3 topics.
 
We also consider the Debate dataset \citep{debate_dataset} as a second independent evaluation data set 
\ma{in \S\ref{sec:systems}}.
Debate  
includes 3,228 argumentative text sequences filtered from posts on four different topics in an online debate forum. The text sequences are labeled with their reason within the respective topic and whether they are supporting (1) or opposing (-1). 
We consider the argumentative text sequences as arguments and the reasons as argument summaries. In contrast to ArgKP21, the dataset exclusively contains matching pairs.  
Exemplary data points for ArgKP21 and Debate are presented in Table~\ref{tab:argkp} and Table~\ref{tab:debate} in the appendix, respectively.

\paragraph{LLMs} 
\manew{We consider four LLMs for the ArgSum systems}
as described in 
\S\ref{sec:ExtractiveSystems}\manew{,} 
\S\ref{sec:Clustering-based_Systems} \manew{and \S\ref{sec:end-to-end}}\yc{, including GPT-4o,\footnote{\url{https://platform.openai.com/docs/models/gpt-4o}} LLaMa3.3-70b \citep{grattafiori2024llama3herdmodels}, Qwen2.5-72b \citep{qwen2025qwen25technicalreport} and Qwen3-32B (non-thinking mode) \citep{yang2025qwen3technicalreport}}. 
GPT-4o-mini\footnote{\url{https://platform.openai.com/docs/models/gpt-4o-mini}} was integrated into the LLM-based evaluation metric as discussed in 
\S\ref{sec:eval}\yc{, since it offers fast response times and is a cost-effective model version for the more quantity-based evaluation approach.} 
\yc{We accessed OpenAI models via the official API\footnote{\url{https://openai.com/index/openai-api/}} and open-source models via the OpenRouter API.\footnote{\url{https://openrouter.ai/}}}

\subsection{Reliability of automatic metrics}\label{sec:metrics} 
To measure the quality of diverse automatic metrics, we correlate them with our human assessment of 126 argument summaries in terms of coverage and redundancy (see \S\ref{sec:eval}). 
For the Soft-Score and CS, we focus exclusively on coverage, as their conceptual design does not allow for any meaningful correlation with redundancy.

We consider two ways of computing correlations. (i) We calculate correlations \textit{across} all topics and stances simultaneously. (ii) We calculate correlations \textit{within} topics and stances and average the results. For the latter scenario, we also report the standard deviations, indicating the variability of reliability.

\paragraph{Soft-Score} We apply the Soft-Score, explained in \S\ref{sec:eval}, with 
the following automatic metrics as similarity function $f$:
(1) ROUGE 1 \citep{lin-2004-rouge}, %1/2/L, 
(2) BERTScore F1 \citep{bertscore}, (3) MoverScore \citep{zhao-etal-2019-moverscore}, (4) BARTScore \citep{bartscore}, (5) BLEURT \citep{sellam-etal-2020-bleurt} and (6) MENLI \citep{chen-eger-2023-menli}. 
Table~\ref{tab:soft_score_corr_cov} in the appendix shows the results. 

\se{First, we note that sP does not intuitively correspond to the 
annotation dimensions of coverage or redundancy in our human annotation --- sP could be interpreted as the fraction of candidate summaries covered by the reference summaries, but not vice versa. Thus, it comes as no surprise that the correlation between sP and coverage is close to zero across all settings.} 
The sR, which better matches the definition of coverage, performs clearly better, even if no strong correlations are observed. Across topics and stances, 
MENLI performs best (0.265) 
followed by BERTScore-F1 (0.254). 
The scenario within topics and stances generally yields better correlation results for the sR. While BERTScore-F1
exhibits the highest correlation at 0.402, 
MENLI (0.372) also achieves a moderate positive correlation with the human coverage scores.  
It is notable that BLEURT and BARTScore, 
suggested by \citet{li-etal-2023-hear}, achieve the poorest results among all considered similarity functions.\footnote{We rescaled BARTScore according to \citet{li-etal-2023-hear} in order to obtain positive scores in the range from zero to one.}

%%%%%%%%%%%%%%%%%%%%%%%%%%%%%%%%%%%%%%%%%
%%%%%%%%%%%%%%%%%%%%%%%%%%%%%%%%%%%%%%%%%%%%
%%%%%%%%%%%%%%%%%%%%%%%%%%%%%%%%%%%%%%%%%%%%

\paragraph{Coverage Score}
To examine the correlation of the CS with the averaged human coverage scores, we consider the Match Scorers of BarH, as proposed by \citet{khosravani-etal-2024-enhancing}, as well as those of SMatchToPr and MCArgSum. Furthermore, we apply various values for the threshold $t_M$, which determines the match score for which an individual reference summary is understood as covered or not. 

As depicted in Table~\ref{tab:coverage_score_corr} in the appendix, the CS 
with BarH's Match Scorer reaches a maximum correlation of 0.489 across and 0.698 within topics and stances. For the scenario across topics and stances, SMatchToPr performs even better and achieves a maximum correlation of 0.541. Within topics and stances, SMatchToPr reaches a maximum correlation of 0.6. The Match Scorer included in MCArgSum yields comparatively worse results, achieving a maximum correlation of 0.449 within and 0.551 across topics and stances. Regarding the matching threshold $t_{M}$, BarH\se{'s Match Scorer} performs very stably across the considered parameter range, whereas this is not the case for both other variants. 

To summarize, the CS provides considerably stronger correlations for the dimension of coverage compared to the Soft-Score.

\paragraph{LLM-based metric} 
The LLM-based metrics for coverage and redundancy,  
described in  
\S\ref{sec:eval}, 
are  examined regarding their respective criterion. 
Here, we investigate different values for the temperature, a parameter controlling the creativity or randomness in LLM-based text generation \citep{peeperkorntemperature}. 
The results are collected in Table~\ref{tab:llm_cov_red_corr}.

The LLM-based score for coverage achieves a maximum correlation of 0.767 across and 0.803 within topics and stances. Consequently, it performs better than the Soft-Score and CS in all scenarios. The LLM-based metric for redundancy reaches also high correlations with a maximum value of 0.852 across and 0.824 within topics and stances. 
Thus, we exclusively use the LLM-based evaluation metrics to assess the argument summarization capability of ArgSum systems in \S\ref{sec:systems}.

\subsection{System evaluation}\label{sec:systems}

\input{results_tables/weighted_new}

Having identified the LLM-based evaluation metrics as the most reliable among those considered for both dimensions of coverage and redundancy, this section addresses their application in order to evaluate the ArgSum systems. 
In our investigations, we make use of a \textit{Weighted Score} assessing both coverage and redundancy simultaneously.
The Weighted Score $ws$ for a certain set of argument summaries is defined as follows:
\begin{equation}
    ws = \alpha \cdot c + (1-\alpha) \cdot (1-r) 
\end{equation}
where $c$ indicates the LLM-based Coverage Score and $r$ indicates the LLM-based Redundancy
Score.  
The weighting factor $\alpha$ is defined to be in the range [0, 1] and can be used to bias the Weighted Score either towards coverage or redundancy. 
For 
our investigations, we set the weighting factor to $2/3$, as we consider coverage to be more important than redundancy.
We generate several argument summaries using various hyperparameter settings (see Appendix \ref{sec:Details on Hyperparameters}) and report the best setting in terms of the Weighted Score for each ArgSum system. 
Since ArgSum is performed per topic and stance, the final evaluation score for each ArgSum system results as the average of the highest Weighted Scores within topics and stances. For simplicity, we refer to the averaged highest Weighted Score as the Weighted Score and the averaged LLM-based Coverage and Redundancy Score as the Coverage and Redundancy Score, respectively. 

\paragraph{Results} 
The Weighted Scores of ArgSum systems are depicted in
Table~\ref{tab:weighted}; we refer to Table~\ref{tab:sum} in Appendix~\ref{sec:Tables and Figures} for full evaluation results including Coverage and Redundancy Scores. 
For both \textbf{classification-based systems}, integrating an LLM generally improves performance. 
\manew{On \ArgKPto{}, 4 out of 8 configurations of \BarH{} and \SMatchToPr{} with LLMs outperform their original versions in Weighted Scores. On Debate, all LLM-enhanced systems achieve higher Weighted Scores than their non-LLM counterparts.} Qwen3-32B is the most effective LLM: it consistently boosts 
\manew{the original} systems and achieves the highest Weighted Scores across all LMM-based variants. 
GPT-4o performs slightly below Qwen-3-32B\manew{, also improving scores in all classification-based cases.} 
While Qwen-2.5-72B yields the lowest performance among all BarH models with LLM integration (0.829 on ArgKP21 and 0.807 on Debate), LLaMA-3.3-70B performs worst among all SMatchToPr variants (0.816 on ArgKP21 and 0.815 on Debate).

\yc{As for \textbf{clustering-based systems}, all \MCArgSum{} variants except those with LLaMA-3.3-70B outperform \USKPM{} in Weighted Scores (0.844-0.853 vs.\ 0.8 on \ArgKPto{}; 0.880-0.898 vs.\ 0.833 on Debate). Similar to the classification-based systems, Qwen-3-32B performs best, ranking first on both datasets, while LLaMA-3.3-70B ranks last on both.
}

\manew{LLaMA-3.3-70B also provides the worst results in \textbf{LLM end-to-end summarization} 
with a Weighted Score of 0.804 on \ArgKPto{} and 0.790 on Debate. As in the other cases of LLM integration, Qwen-3-32B achieves the best results with a Weighted Score of 0.923 on \ArgKPto{} and 0.899 on Debate, followed by Qwen-2.5-72B (0.904 and 0.858) and GPT-4o (0.856 and 0.841).}  

Qualitative inspection of the different systems' outputs (generated summaries) shows that the low scores of LLaMA-3.3-70B 
%are to 
\ycr{may} be attributed to the model's tendency to create very short summaries (bullet point style), while Qwen-3-32B and GPT-4o mostly produce full sentences (consistently across datasets). 
E.g., on the topic of ``The USA is a good country to live in'', LLaMA-3.3-70B creates summaries like ``Offers freedom'' or ``Has many freedoms'', while Qwen-3-32B produces summaries on the same topic like ``The USA offers unparalleled freedom and the American dream.'' or ``High levels of freedom and democratic values.'' 
\ycr{We refer to Table~\ref{tab:ex} in the appendix for more examples.}
We did not notice any systematic differences in the systems' outputs across topics and/or stances. 

Overall, the integration of LLMs results in considerable improvements for classification-based as well as clustering-based ArgSum systems. \yc{On \ArgKPto{}, classification-based systems outperform clustering-based ones on average, particularly those using LLM-based candidate generation, whereas both system types perform comparably on Debate.}
\manew{The fact that LLM end-to-end summarization with Qwen3-32B yields the best overall results on both datasets raises the question of whether traditional \ArgSum{} systems, \sen{leveraging different components such as Match and Quality Scorers or clustering techniques}, should still be used at all.}

The final choice of an \ArgSum{} system should also depend on the runtime requirements.  
\yc{Using GPT-4o as an example}, clustering-based systems are generally faster, with MCArgSum showing \manew{suitable} 
performance  
for both datasets. It required on average 3.78 seconds per topic and stance for ArgKP21 and 7.77 seconds for Debate (cf. hardware specifications in Appendix~\ref{subsec:Hardware}).\footnote{BarH+cand(gpt-4o) required on average 36.57 seconds on ArgKP21 and 65.34 seconds on Debate, whereas the LLM end-to-end summarization with GPT-4o took on average 8.37 seconds and 14.86 seconds, respectively.}

%% file: results_tables/coverage.tex
\begin{table*}[!t]
\small
\centering
\resizebox{\textwidth}{!}{
\begin{tblr}{
  colspec = {X[c,m]||X[c,m]|X[c,m]|X[c,m]||X[c,m]|X[c,m]|X[c,m]},
  width=1\textwidth,
  stretch = 0,
  rowsep = 4pt,
}
\hline
\hline
\SetCell[r=2]{c} {Temper-\\ature} & \SetCell[c=3]{c} \textbf{LLM-based Coverage Score} & & 
 & \SetCell[c=3]{c} \textbf{LLM-based Redundancy Score} & & \\
& Across & Within & Runtime (s) & Across & Within & Runtime (s) \\
\hline
0.20 & 0.736 & {0.756 {$\pm$ 0.100}} & 495.1 & 0.798 & {0.697 {$\pm$ 0.226}} & 1305.3\\
\hline[dotted]
0.30 & 0.725 & {0.747 {$\pm$ 0.133}} & 467.7 & 0.789 & {0.752 {$\pm$ 0.096}} & 1651.1 \\
\hline[dotted]
0.40 & 0.746 & {0.771 {$\pm$ 0.112}} & 529.3 & 0.817 & {0.758 {$\pm$ 0.122}} & 1515.4 \\
\hline[dotted]
0.50 & 0.742 & {0.757 {$\pm$ 0.127}} & 512.0 & 0.812 & {0.724 {$\pm$ 0.210}} & 1446.3 \\
\hline[dotted]
0.60 & 0.741 & {0.762 {$\pm$ 0.122}} & 629.7 & 0.837 & {0.795 {$\pm$ 0.088}} & 1359.9 \\
\hline[dotted]
0.70 & 0.755 & {0.789 {$\pm$ 0.103}} & 644.3 & 0.830 & {0.782 {$\pm$ 0.112}} & 1425.6 \\
\hline[dotted]
0.80 & 0.729 & {0.755 {$\pm$ 0.108}} & 612.4 & 0.828 & {0.762 {$\pm$ 0.111}} & 1431.1 \\
\hline[dotted]
0.90 & 0.754 & {0.782 {$\pm$ 0.131}} & 676.4 & 0.843 & {0.784 {$\pm$ 0.109}} & 1651.0 \\
\hline[dotted]
1.00 & \textbf{0.767} & {\textbf{0.803} {$\pm$ 0.115}} & 845.5 & \textbf{0.852} & {\textbf{0.824} {$\pm$ 0.055}} & 1649.1 \\
\hline
\hline
\end{tblr}
}
\caption[Correlation between LLM-based Coverage and Redundancy Scores and the respective averaged human scores.]{Pearson correlation between the LLM-based Coverage and Redundancy Scores and the respective averaged human scores for different temperatures, along with the evaluation runtime, \se{on ArgKP21}. For the scenario within topics and stances, %as detailed in \autoref{Experimental Setup soft-score}, 
standard deviations are indicated alongside the correlation values. 
%The strongestpositive correlations for both scenarios within each LLM-based score are highlighted in bold.
}
\label{tab:llm_cov_red_corr}
\vspace{-.5cm}
\end{table*}

%% file: results_tables/weighted_new.tex
\begin{table}[!t]
\small
\resizebox{\columnwidth}{!}{%
\begin{tabular}{@{}ll|l@{}}
\toprule
                          & ArgKP21                         & Debate       \\ \midrule
\multicolumn{3}{l}{\emph{Classification-based}   }                                                                          \\ \midrule
BarH                           & \multicolumn{1}{l|}{0.848}    &      0.770    \\
BarH+cand(gpt-4o)               & \multicolumn{1}{l|}{0.877 $\uparrow$}    &  0.847 $\uparrow$   \\
BarH+cand(llama-3.3-70b)       & \multicolumn{1}{l|}{0.845}       & 0.807 $\uparrow$   \\
BarH+cand(qwen-2.5-72b)       & \multicolumn{1}{l|}{0.829}    &  0.807 $\uparrow$   \\
BarH+cand(qwen3-32b)          & \multicolumn{1}{l|}{0.900 $\uparrow$}    &  0.880 $\uparrow$   \\ \midrule
SMtPR                          & \multicolumn{1}{l|}{0.856}      & 0.805    \\
SMtPR+cand(gpt-4o)           & \multicolumn{1}{l|}{0.884 $\uparrow$}    & 0.869 $\uparrow$   \\
SMtPR+cand(llama-3.3-70b)       & \multicolumn{1}{l|}{0.816}       & 0.815 $\uparrow$   \\
SMtPR+cand(qwen-2.5-72b)       & \multicolumn{1}{l|}{0.843}    & 0.860 $\uparrow$   \\
SMtPR+cand(qwen3-32b)         & \multicolumn{1}{l|}{0.896 $\uparrow$}     & 0.890 $\uparrow$   \\ \midrule
\multicolumn{3}{l}{\emph{Clustering-based}    }                                                                             \\ \midrule
USKPM                        & \multicolumn{1}{l|}{0.800}     & 0.833    \\
MCArgSum(gpt-4o)              & \multicolumn{1}{l|}{0.844}     & 0.886    \\
MCArgSum(llama-3.3-70b)     & \multicolumn{1}{l|}{0.765}      & 0.729    \\
MCArgSum(qwen-2.5-72b)      & \multicolumn{1}{l|}{0.847}    & 0.880    \\
MCArgSum(qwen3-32b)           & \multicolumn{1}{l|}{0.853}       & 0.898    \\ 
\midrule
\multicolumn{3}{l}{\emph{LLM end-to-end Summarization}    }                                                                             \\ \midrule
gpt-4o              & \multicolumn{1}{l|}{0.856}     & 0.841    \\
llama-3.3-70b     & \multicolumn{1}{l|}{0.804}      & 0.790    \\
qwen-2.5-72b      & \multicolumn{1}{l|}{0.904}    & 0.858    \\
qwen3-32b           & \multicolumn{1}{l|}{\textbf{0.923}}       & \textbf{0.899}    \\ \bottomrule
\end{tabular}}
\caption{\textbf{Weighted Scores} of ArgSum systems on ArgKP21 and Debate datasets. For BarH and SMatchToPr (abbreviated as SMtPR), the variant with LLM-based candidates is indicated by +cand and the models in brackets indicate the LLMs integrated. We bold the best results on each dataset. $\uparrow$ indicates that classification-based systems with LLM integration outperform the original systems.}\label{tab:weighted}
\vspace{-.5cm}
\end{table}

%\todo{YC-old: bold and underline some results}

%% file: sections/6_conclusions_may.tex
\section{Conclusion}

 Our proposed LLM-based ArgSum systems and metrics achieve state-of-the-art performance across the two datasets considered. 
 \MCArgSum{} and the LLM end-to-end summarization, our newly proposed LLM-based ArgSum systems,  \yc{outperform existing approaches.} 
 \ycr{LLM end-to-end summarization achieves the best performance on both datasets, while \MCArgSum{} ranks second on debate but is generally faster.}
 \ycr{Among all LLMs tested, Qwen-3-32B benefits ArgSum systems the most.}
 The LLM-based ArgSum evaluation scores we propose show very high correlation with human judgments and thus set a very reliable evaluation framework where reference summaries are available.

 A few open questions and tasks remain: 
 \yc{While we applied uniform prompts and parameter settings across all LLMs for consistency, optimizing them for each model may unlock further performance gains. }
 Furthermore, we leave the application of reference-free evaluation strategies to future work.

%% file: sections/7_ack_limit_ethic.tex
\section*{Limitations} \label{sec:limit}
\yc{All inspected LLMs were trained on data that postdates the publication of \citet{debate_dataset} and \citet{friedman-etal-2021-overview}. Therefore, the evaluation datasets used in this work may have been seen during their \sen{(pre-)}training.
}
\se{However, similar limitations of potential data contamination are faced in many other recent problem settings as well; due to a lack of suitable ArgSum datasets, this issue is hard to avoid. We also point out that this work introduces a new evaluation benchmark for ArgSum systems, which could not have been seen by our employed LLMs.} 
\yc{Additionally, prompts were initially designed for GPT-4o and applied uniformly across all LLMs, which may have resulted in an overestimation of GPT-4o's performance.
 Nevertheless, some LLMs still outperform GPT-4o in our evaluation.
The system outputs included in the human evaluation do not cover those from \ArgSum{} systems using the open-source LLMs. 
}

\section*{Ethical Considerations} \label{sec:ethical}
ArgSum systems could yield unreliable, factually incorrect, biased or even maliciously misleading summaries of the underlying source arguments --- particularly, if certain arguments are misrepresented or filtered.
Thus, the usage of ArgSum systems must always be made transparent, and recipients of the summarized arguments must interpret these with care. 

We used ChatGPT solely for text refinement during the writing of this paper.

%% file: sections/100_appendix.tex
\appendix

\section{Details on Model Fine-tuning}\label{sec:Details on Model Fine-tuning}

\paragraph{BarH and SMatchToPr} We fine-tuned the Match Scorers and Quality Scorers in BarH and SMatchToPr according to \citet{bar-haim-etal-2020-quantitative} and \citet{alshomary-etal-2021-key}, respectively. It is important to note that \citet{bar-haim-etal-2020-quantitative} do not specify which of the two quality scores (MACE-P and WA) in ArgQ should be used for training the Quality Scorer. Additionally, it is unclear whether a model with or without a pooling layer was used. Since the model without pooling layer and fine-tuned on MACE-P performs best in preliminary investigations, we applied it in BarH. 

\paragraph{USKPM} The fine-tuning of FLAN-T5 in USKPM was conducted as proposed by \citet{li-etal-2023-hear}, though no specific learning rate was provided. Based on our observations, a learning rate of 4e-4 worked well and was therefore used for fine-tuning the model.

\paragraph{MCArgSum} As Match Scorer, MCArgSum uses the SBERT model ``all-mpnet-base-v2'' fine-tuned on ArgKP21. The fine-tuning is conducted over 10 epochs with a learning rate of 5e-6 and contrastive loss. The best performing model on the development set was selected as final model.

\section{LLM Prompting}\label{sec:LLM Prompting}

LLM prompting can be divided into a system message and a user message. The system message guides the LLM on its general behavior, while the user message specifies the exact task. We also utilize the system message to introduce the task at hand and to describe the desired appearance of the argument summaries.

\subsection{Classification-based Systems \manew{and LLM end-to-end Summarization}}\label{subsec:LLM Prompting - Classification-based Systems}

The proposed prompting strategy instructs the LLM to generate either candidates or \manew{directly generate} argument summaries. In both cases, the prompt is divided into a system message and a user message. The following prompt template is applied for both generating candidates and argument summaries, but used differently. For generating candidates, we instruct the LLM to produce a large number of key points (12 to 20). In contrast, for argument summaries, we request fewer key points (4 to 8) and apply the 
\manew{additional} user message to minimize redundancy. A description of the parameters and placeholders contained in the prompt template is given below.

\paragraph{System Message} You are a professional debater and you can express yourself succinctly. If you are given a corpus of arguments on a certain debate topic and stance, you find \texttt{\{num\_kps\}} appropriate salient single sentences, called key points, summarizing most of the  arguments and providing a textual and quantitative view of the data. A key point can be seen as a meta argument why one is for or against a certain topic. Make sure that the generated key points summarize the majority of the arguments contained in the corpus. A key point should not exceed a length of \texttt{\{kp\_token\_length\}} tokens. Here are two examples of good key points: ``School uniform reduces bullying'' is an opposing key point on the topic ``We should 
abandon the use of school uniform" and "Guns lead to accidental deaths'' is a supporting key point on the topic "We should abolish the right to keep and bear arms".

\paragraph{User Message} Please generate \texttt{\{num\_kps\}} short (maximal length of \texttt{\{kp\_token\_length\}} tokens), salient and high quality \texttt{\{stance\}} key points on the topic ``{\{\texttt{topic}\}}'' so that they capture the main statements that are shared between most of the arguments based on the following corpus of arguments: \texttt{\{arguments\}}.

\paragraph{
\manew{Additional} User Message for 
\manew{LLM end-to-end Summarization}} You should only generate as many key points as necessary to summarize the arguments contained in the corpus. This means you should 
preferably generate fewer key points than the maximum permitted number of \texttt{\{max\_num\_kps\}} key points instead of generating overlapping key points in terms of content.

\paragraph{Parameters/Placeholders}
\begin{itemize}[topsep = 0mm, itemsep = -0.75mm]
    \item \texttt{num\_kps}: Number of key points (can be a fixed value or a range of values)
    \item \texttt{kp\_token\_length}: Maximum permitted number of tokens for key points
    \item \texttt{stance}: Stance of arguments (supporting or opposing)
    \item \texttt{topic}: Topic of arguments
    \item \texttt{arguments}: List of arguments
    \item \texttt{max\_num\_kps}: Maximum permitted number of key points
\end{itemize}

\subsection{Clustering-based Systems}\label{subsec:LLM Prompting - Clustering-based Systems}

The prompting for LLM-based Cluster Summarization is divided into a system message and a user message. A description of the parameters and placeholders contained in the prompt template is given below.

\paragraph{System Message} You are a professional debater and you can express yourself succinctly. If you are given a cluster of similar arguments on a certain debate topic and stance, you find a single appropriate salient sentences, called key point, capturing the main statement that is shared between most of the clustered arguments and providing a textual and quantitative view of the data. A key point can be seen as a meta argument why one is for or against a certain topic. Since argument clusters are not perfect, they may contain arguments that do not actually belong together. Therefore, make sure that a generated key point summarizes the majority of the arguments contained in the cluster. A key point should not exceed a length of \texttt{\{kp\_token\_length\}} tokens. Here are two examples of good key points: ``School uniform reduces bullying'' is an opposing key point on the topic ``We should abandon the use of school uniform" and ``Guns lead to accidental deaths'' is a supporting key point on the topic ``We should abolish the right to keep and bear arms''. 

\paragraph{User Message} Please generate a single short (maximal length of \texttt{\{kp\_token\_length\}} tokens), salient and high quality \texttt{\{stance\}}
key point on the topic ``\texttt{\{topic\}}'' so that it captures the main statement that is shared among most of the clustered arguments 
for each of the following \texttt{\{num\_clusters\}} clusters of similar arguments: \texttt{\{clusters\}}. Since argument clusters are not perfect, they may contain arguments that do not actually belong together. Therefore, make sure that each generated key point summarizes the majority of the arguments contained in the respective cluster. In addition, ensure that 
the generated key points do not overlap in terms of content. Do not deliver an explanation why you generated the key points or any other information. Only return the cluster ids and corresponding individual key points. 

\paragraph{Parameters/Placeholders}
\begin{itemize}[topsep = 0mm, itemsep = -0.75mm]
    \item \texttt{kp\_token\_length}: Maximum permitted number of tokens for key points
    \item \texttt{stance}: Stance of arguments (supporting or opposing)
    \item \texttt{topic}: Topic of arguments
    \item \texttt{arguments}: List of arguments
    \item \texttt{num\_clusters}: Number of clusters
    \item \texttt{clusters}: List of argument clusters, where each cluster consists of a cluster id and a list of the corresponding arguments
\end{itemize}

\subsection{LLM-based Evaluation}

For the evaluation, we only worked with user messages.
 
\paragraph{User Message for Coverage Evaluation}

Your task is to evaluate a set of generated summaries obtained from a collection of arguments against a set of reference summaries. The evaluation is conducted according to the criteria of coverage, meaning that the set of generated summaries aims to cover the main statements contained in the set of reference summaries. Since each reference summary addresses a unique main statement, you are asked to count the number of reference summaries that are covered by the set of generated summaries. If a reference summary is only partially covered by the set of generated summaries, an increase of the count by 0.5 is allowed. Your counts aim to correlate well with human judgments.  
 
Make sure to always print the final count in the format "Coverage count: x.y" in a new line with no additional text in that line.\vspace{1mm}\\
Example:\vspace{1mm}\\
\textbf{Set of Reference Summaries}:
\begin{enumerate}[topsep = 1mm, itemsep = -0.75mm, leftmargin = 10mm]
    \item Banning guns would save lives
    \item Guns can fall into the wrong hands 
    \item Guns lead to accidental deaths
    \item Gun ownership allows for mass-shootings/general gun violence
\end{enumerate}
\textbf{Set of Generated Summaries:} 
\begin{enumerate}[topsep = 1mm, itemsep = -0.75mm, leftmargin = 10mm]
    \item Banning guns would save thousands of lives
    \item Some people do not know how to handle firearms. This is a danger to them and others.  
    \item Guns kill people, they should be banned  
    \item Firearms can fall into the hands of potential murderers 
    \item Firearms are a disgrace to humanity.  
    \item Without weapons, there would be no war. 
\end{enumerate}
Coverage count: 3.5
\vspace{1mm}\\
\textbf{Evaluation Procedure}:
\begin{enumerate}[topsep = 1mm, itemsep = -0.75mm, leftmargin = 10mm]
    \item Read the reference summaries. Do not print them again.
    \item Read the generated summaries. Do not print them again.
    \item Go through the set of reference summaries and determine whether the reference summary at hand is covered by at least one generated summary.
    \item Once you have done this for each reference summary, count the number of covered reference summaries and return the resulting coverage count.
\end{enumerate}
\vspace{1mm}
\textbf{Evaluation Task:}
\newline
Set of Reference Summaries:\vspace{1mm}\\
\texttt{reference\_summaries}
\vspace{1mm}\\
Set of Generated Summaries:\vspace{1mm}\\
\texttt{candidate\_summaries}
 
\paragraph{User Message for Redundancy Evaluation} 

Your task is to evaluate a set of arguments on a certain debate topic and stance according to their uniqueness. Since arguments can be formulated differently, but address the same aspect of a debate, your task is to count the number of unique main statements addressed by the set of arguments. If a main statement addressed by an argument is only partially unique because it is also in parts covered by another argument, an increase of the count by 0.5 is allowed. Your counts aim to correlate well with human judgments.  
 
In the following, you are provided with an example, instructions for the evaluation procedure, and finally with your evaluation task.
\vspace{1mm}\\
Example:\vspace{1mm}\\ 
\textbf{Set of Arguments:}
\begin{enumerate}[topsep = 1mm, itemsep = -0.75mm, leftmargin = 10mm]
    \item Banning guns would save lives  
    \item Guns can fall into the wrong hands  
    \item Guns lead to accidental deaths  
    \item Guns kill people, they should be banned  
    \item Gun ownership allows for mass-shootings/general gun violence \item Some people do not know how to handle firearms. This is a danger to them and others.  
    \item Banning guns would save thousands of lives  
    \item Firearms can fall into the hands of potential murderers  
\end{enumerate}
\vspace{1mm}
Number of Unique Main Statements: 4
\vspace{1mm}\\
\textbf{Explanation:}
\begin{itemize}[topsep = 1mm, itemsep = -0.75mm, leftmargin = 10mm]
\item  Argument 1, 4, and 7 address the same main statement (guns kill people so without guns lives could be saved)  
\item  Argument 2, 6, and 8 address the same main statement (guns could fall into the wrong hands, such as murders or people not knowing how to handle guns)  
\item Argument 3 addresses a unique main statement, focusing on accidents with guns  
\item  Argument 5 addresses a unique main statement, focusing on intentional killing like terrorism or running amok  
\end{itemize}
\vspace{1mm}
\textbf{Notes:}
\begin{itemize}[topsep = 1mm, itemsep = -0.75mm, leftmargin = 10mm]
\item Arguments 1, 4, and 7 are quite general, and therefore differ from the others  
\item E.g., argument 3 could also be assigned to 1, 4, and 7. Nevertheless, it focuses on accidents and is more specific 
\end{itemize}
\vspace{1mm}
\textbf{Evaluation Procedure:}
\begin{enumerate}[topsep = 1mm, itemsep = -0.75mm, leftmargin = 10mm]
    \item Read the arguments. Do not print them.
    \item Go through the list of arguments, starting with the first argument.
    \item Determine whether the argument at hand addresses a main statement of the debate.
    \item Move on to the next one and consider whether it addresses a main statement and whether it has already been covered by previous arguments in the list.
    \item Once you have done this for each argument, count the total number of unique main statements.
    \item Return your uniqueness count in the format "Number of Unique Main Statements: x.y" in a new line with no additional text in that line. Always make this line the last line of your response and always include it.
\end{enumerate}
\vspace{1mm}
\textbf{Evaluation Task:}
\vspace{1mm}\\
\textbf{Set of Arguments:}
\texttt{candidate\_summaries}
\vspace{1mm}\\
Number of Unique Main Statements:

\paragraph{Generation of Candidate Summaries and Reference Summaries}
 
Candidate Summaries and Reference Summaries are constructed by iterating over lists of generated and reference summaries, respectively. Each element in the list is formatted as an enumerated string, where each entry is prefixed with its index and a period. This ensures a structured representation of arguments or summaries for evaluation. Below is an example of how a list of three reference summaries would be converted into a formatted string:
\vspace{1mm}\\
\textbf{Set of Reference Summaries:}
\begin{enumerate}[topsep = 1mm, itemsep = -0.75mm, leftmargin = 10mm]
\item Renewable energy reduces carbon emissions.
\item Solar panels provide long-term cost savings.
\item Wind power is a reliable energy source.
\end{enumerate}
\vspace{1mm}
Similarly, a set of generated summaries follows the same structure, ensuring consistency in comparison.

\section{Human Evaluation}\label{sec:Human Evaluation}

\subsection{Introduction to ArgSum}\label{subsec: Introduction to ArgSum}
A debate on a certain topic can be conducted using a variety of arguments for each side of the debate. Although some of these arguments refer to the same main statement, they can be formulated very differently. While the number of possible arguments seems to be almost infinite due to the possibility of different formulations, the number of possible main statements within a debate is limited.

Argument summarization is about summarizing a relatively large set of arguments on a certain debate topic and stance by generating a small set of argument summaries, each expressing one distinct main statement contained in the set of arguments. In addition, each argument is matched to the generated summary that conveys its main statement the best. Following is a simple example: \\

\begingroup
\leftskip = 0mm 
\noindent\textbf{Topic}: \vspace{1mm}\\
We should abandon the use of school uniform \vspace{1mm} \\
\textbf{Stance}: \vspace{1mm}\\
Opposing \vspace{1mm}\\
\textbf{Set of Arguments}: \vspace{1mm}
\begin{enumerate}[topsep = 1mm, itemsep = -0.75mm, leftmargin = 10mm]
% was 300
    \item \ctext[RGB]{255,209,82}{School uniforms keep everyone looking the same and prevent bullying} 
    \item \ctext[RGB]{80,230,100}{School uniforms can help parents save money on outfit}
% was 300
    \item \ctext[RGB]{255,209,82}{School uniforms help stop bullying because when people are similarly dressed, nobody is made to feel inferior}
    \item \ctext[RGB]{80,230,100}{It is cheaper for parents to buy school uniforms, which is helpful to parents that are struggling financially}
    \item \ctext[RGB]{80,230,100}{School uniforms are substantially more affordable}
\end{enumerate}
\vspace{1mm}
\textbf{Set of Summaries}: \vspace{1mm}
\begin{enumerate}[topsep = 1mm, itemsep = -0.75mm, leftmargin = 10mm]
% was 300
    \item \ctext[RGB]{255,209,82}{School uniforms reduce bullying}
    \item \ctext[RGB]{80,230,100}{School uniforms save costs}
\end{enumerate}
\vspace{1mm}
\textbf{Argument Summary Matches}: \vspace{1mm} \\
The matches are highlighted by the colored markings:
\begin{itemize}[topsep = 1mm, itemsep = -0.75mm, leftmargin = 10mm]
    \item Arguments 1 and 3 are matched to summary 1
    \item Arguments 2, 4 and 5 are matched to summary 2
\end{itemize}
\endgroup

\subsection{Description of the Evaluation Task}\label{subsec:Description of the Evaluation Task}

This task is about determining how well a set of generated argument summaries serves as a summary of possible arguments on a certain debate topic and stance. 

For this purpose, you are given a set of generated summaries and a set of reference summaries as well as the corresponding debate topic and stance. You have to carry out the following two instructions regarding the dimensions of coverage and uniqueness:

\begin{enumerate}[topsep = 2.5mm, itemsep = -0.75mm, leftmargin = 10mm]
    \item \textbf{Coverage}: Count the number of reference summaries that are covered by the set of generated summaries. 
    \item \textbf{Uniqueness}: Count the number of distinct/unique main statements contained in the set of generated summaries. 
\end{enumerate}

For both dimensions increments of 0.5 are allowed. In the case of coverage, this applies if a reference summary is only partially covered by the set of generated summaries. For the dimension of redundancy, this applies if there is a distinct main statement in the set of generated summaries that partially overlaps with another. For the case you are not sure, you can answer with -1. Following is an example:\\

\begingroup
\leftskip = 0mm 
\noindent\textbf{Topic}: \vspace{1mm}\\
Routine child vaccinations should be mandatory \vspace{1mm}\\
\textbf{Stance}: \vspace{1mm}\\
Opposing \vspace{1mm}\\
\textbf{Set of Reference Summaries}: \vspace{1mm}
\begin{enumerate}[topsep = 1mm, itemsep = -0.75mm, leftmargin = 10mm]
    \item Mandatory vaccination contradicts basic rights
    \item Routine child vaccinations are not necessary to keep children healthy
    \item Routine child vaccinations, or their side effects, are dangerous
    \item The parents and not the state should decide
\end{enumerate}
\vspace{1mm}
\textbf{Set of Generated Summaries}: \vspace{1mm}
\begin{enumerate}[topsep = 1mm, itemsep = -0.75mm, leftmargin = 10mm]
    \item Vaccinations violate free will and personal choice
    \item Mandatory vaccines conflict with religious beliefs
    \item Parents should have the right to decide
    \item Children may suffer harmful effects from vaccines
    \item Concerns about vaccine safety and side effects
\end{enumerate}
\vspace{1mm}
\textbf{Coverage}: \vspace{1mm}\\
3 (The second reference summary is not covered.) \vspace{1.5mm}\\
\textbf{Uniqueness}: \vspace{1mm}\\
3.5 (The first and third generated summaries address two different distinct main statements. The fourth and fifth generated summaries refer to the same distinct main statement. The second generated summary partially overlaps with the first one.)
\endgroup

\section{Experimental conditions}\label{Experimental conditions}

\subsection{Data Preprocessing}\label{sec:preprocessing}
To conduct our investigations on the test split of ArgKP21 as well as Debate, we performed two pre-processing steps.  
First, we remove arguments that do not have exactly one matching argument summary. The reason for this is that we aim to process only those arguments that have a well-defined reference summary. This 
is 
because the considered automatic evaluation metrics are reference-based. Including arguments without any reference could result in candidate summaries that are not captured by the references and thus bias the evaluation of ArgSum systems. 

Second, we exclude  arguments consisting of more than one sentence, as we consider an adequate argument to consist of a single sentence. This is particularly crucial for the argumentative text sequences contained in Debate. For the test split of ArgKP21, the pre-processing reduces the number of arguments from 732 to 428, while for Debate it is reduced from 3180 to 2321. Finally, to decrease the computational effort, we select only 50\% of the arguments for each unique argument summary in Debate as our final dataset. This pre-processing step results in 1165 remaining arguments for Debate, while retaining each unique argument summary. 

\subsection{Modifications to ArgSum Systems} \label{subsec:Modifications to ArgSum Systems}

We had to apply three modifications to the ArgSum systems as proposed in 
\S\ref{sec:setup}. 
The first concerns the candidate selection in BarH and SMatchToPr. In cases where the proportion of candidates out of all arguments is below a certain threshold $p_{C}$, we 
fill this gap with the highest quality arguments 
not yet considered as candidates. In this way, we avoid cases in which no candidates are identified at all, as the Quality Scorer provides low scores across all arguments. Second, when selecting candidates in SMatchToPr, we delete arguments consisting of several sentences instead of separating them. Finally, we use the Quality Scorer included in BarH instead of TextRank for determining the order of arguments in the corresponding input list of Flan-T5 in \gls{USKPM}.

\subsection{Details on Hyperparameters}\label{sec:Details on Hyperparameters}

When applying BarH and SMatchToPr, we used the recommended parameter values from \citet{bar-haim-etal-2020-quantitative} and \citet{alshomary-etal-2021-key}, respectively. In case of USKPM and MCArgSum, we set the minimum cluster size $c$ to 3. The similarity threshold for IC in USKPM was set to zero, meaning that we forced each unclustered argument to be assigned to an existing cluster. In addition, \autoref{tab:Appendix Hyperparameter Clustering-based Systems} includes the varying hyperparameter settings for the argument clustering inherent in USKPM and MCArgSum. For USKPM, we performed the clustering for each possible combination of the depicted parameter values.

\subsection{Hardware} \label{subsec:Hardware}

We conducted our experiments on a personal computer with an Apple M1 Max chip, which is designed as a system-on-a-chip. It includes a 10-core CPU (8 performance cores and 2 efficiency cores), a 32-core GPU, and a 16-core Neural Engine. The GPU has direct access to the entire main memory of 64GB. The system runs on macOS Sonoma 14.1.2 (64-bit). With the introduction of Metal support for PyTorch on macOS, utilizing the GPU for machine learning tasks has become accessible.\footnote{\url{https://pytorch.org/blog/introducing-accelerated-pytorch-training-on-mac}} This setup was used for both training and inference of PyTorch models.

\section{Tables and Figures}\label{sec:Tables and Figures}

\begin{table*}[!h]
\begin{tblr}{
  colspec = {X[c,m,-1]|X[c,m]|X[c,m]|X[c,m]},
  %hline{1} = {2-5}{},
  width=1\textwidth,
  stretch = 0,
  rowsep = 4pt,
  %hlines = {black, 0.5pt},
  %vlines = {black, 0.5pt}
}
\hline
\hline
  & \textbf{Parameter} & \textbf{Value Range} & \textbf{Steps} \\
\hline
\SetCell[r=3]{m} \textbf{USKPM}
 & Reduced embedding dimensionality & $[2,5]$ & $1$ \\
 \hline[dotted]
 & Number of neighboring samples used for the
manifold approximation of UMAP & $[2,5]$ & $1$ \\
 \hline[dotted]
 & Minimum permitted distance of points in the low
dimensional representation of UMAP  & $[0,0.4]$ & $0.2$ \\
\hline[dotted]
\SetCell[r=1]{m} \textbf{MCArgSum}
 & Minimum match score required between two clusters to be merged ($m$) & $[0.05,0.95]$ & $0.025$ \\
\hline
\hline
\end{tblr}
\caption{Hyperparameter settings of clustering-based ArgSum systems considered in our investigations.} 
\label{tab:Appendix Hyperparameter Clustering-based Systems}
\end{table*}

\begin{figure*}[!h]
    \centering
    \includegraphics[width=0.9\textwidth]{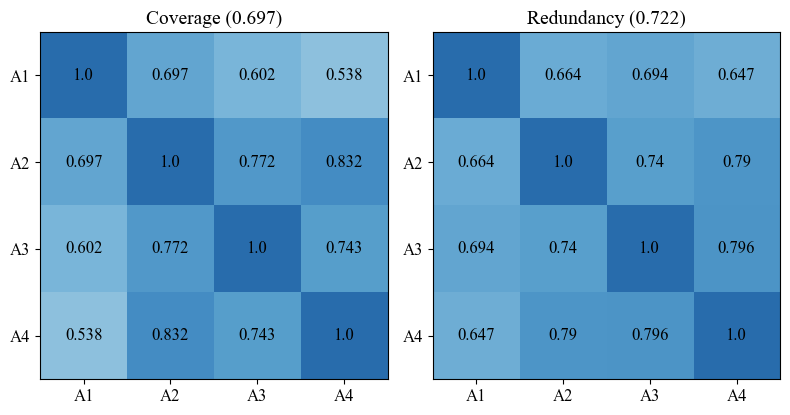}
    \caption[Pairwise correlation of human judgments for the criteria of coverage and redundancy.]{Pairwise Pearson correlation coefficient of the human judgments by the four annotators (A1-A4) for the criteria of coverage and redundancy. The averaged value across annotator pairs is indicated in the parentheses.}
    \label{fig:inter_rater}
\end{figure*}

\begin{table*}[h]
\begin{tblr}{
  colspec = {X[l,m,-1]|X[l,m]},
  width=1\textwidth,
  stretch = 0,
  rowsep = 4pt,
  %hlines = {black, 0.5pt},
  %vlines = {black, 0.5pt}
}
\hline
\hline
\textbf{Topic} & We should abandon the use of school uniform \\
\hline[dotted]
\textbf{Stance} & -1 \\
\hline[dotted]
\textbf{Argument} & school uniforms cut down on bulling and keep everyone the same. \\
\hline[dotted]
\textbf{Key Point} & School uniform reduces bullying \\
\hline[dotted]
\textbf{Label} & 1 \\
\hline[dotted]
\textbf{Set} & dev \\
\hline
\hline
\end{tblr}
\caption[Exemplary data point of ArgKP21.]{Exemplary data point of ArgKP21.}
\label{tab:argkp}
\end{table*}

\begin{table*}[h]
\begin{tblr}{
  colspec = {X[l,m,wd=21mm]|X[l,m]},
  width=1\textwidth,
  stretch = 0,
  rowsep = 4pt,
  %hlines = {black, 0.5pt},
  %vlines = {black, 0.5pt}
}
\hline
\hline
\textbf{Topic} & obama \\
\hline[dotted]
\textbf{Stance} & -1 \\
\hline[dotted]
\textbf{Argument} & Where are those outspoken democrats who voted for him because they were told, no promised, that he would END THE WAR? \\
\hline[dotted]
\textbf{{Argument\\Summary}} & Wars are still on \\
\hline
\hline
\end{tblr}
\caption[Exemplary data point of Debate.]{Exemplary data point of Debate.}
\label{tab:debate}
\end{table*}

\begin{table*}[!t]
\small
\begin{tblr}{
  colspec = {X[c,m, wd=20mm]||X[c,m]|X[c,m]||X[c,m]|X[c,m]||X[c,m]|X[c,m]||X[c,m,wd=17mm]},
  %hline{1} = {2-5}{},
  width=1\textwidth,
  stretch = 0,
  rowsep = 4pt,
  %hlines = {black, 0.5pt},
  %vlines = {black, 0.5pt}
}
\hline
\hline
\SetCell[r=2]{c} {Similarity \\ Function} & \SetCell[c=2]{c} \textbf{Soft-Precision} & 2-2 
 & \SetCell[c=2]{c} \textbf{Soft-Recall} & 2-4
 & \SetCell[c=2]{c} \textbf{Soft-F1} & 2-6 
 & \SetCell[r=2]{c} {Run-\\time (s)} \\
& Across & Within & Across & Within & Across & Within & \\
\hline
{ROUGE 1} & -0.118 & {-0.072 \\ {$\pm$0.194}} & 0.164 & {0.315 \\ {$\pm$0.170}} & 0.027 & {0.127 \\ {$\pm$0.184}} & 0.428 \\
\hline[dotted]
{{BERTSc.} \\ {F1}} & -0.028 & {0.092 \\ {$\pm$0.262}} & 0.254 & {\textbf{0.402} \\ {$\pm$0.175}} & 0.121 & {0.240 \\ {$\pm$0.242}} & 354.2 \\
\hline[dotted]
{MoverSc.} & -0.046 & {0.044 \\ {$\pm$0.227}} & 0.156 & {0.310 \\ {$\pm$0.204}} & 0.069 & {0.191 \\ {$\pm$0.207}} & 55.93 \\
\hline[dotted]
{{BARTSc.} \\ {CNN/DM}} & -0.146 & {-0.305 \\ {$\pm$0.164}} & 0.024 & {-0.011 \\ {$\pm$0.283}} & -0.053 & {-0.132 \\ {$\pm$0.264}} & 84.33 \\
\hline[dotted]
{{BARTSc.} \\ {Parabank}} & -0.271 & {-0.221 \\ {$\pm$0.251}} & -0.012 & {0.112 \\ {$\pm$0.339}} & -0.132 & {-0.022 \\ {$\pm$0.320}} & 41.09 \\
\hline[dotted]
{BLEURT}  & -0.209 & {-0.218 \\ {$\pm$0.289}} & 0.033 & {0.138 \\ {$\pm$0.247}} & -0.091 & {-0.055 \\ {$\pm$0.294}} & 487.3 \\
\hline[dotted]
{MENLI} & -0.154 & {-0.039 \\ {$\pm$0.287}} & \textbf{0.265} & {0.372 \\ {$\pm$0.260}} & 0.107 & {0.228 \\ {$\pm$0.298}} & 254.6 \\
\hline
\hline
\end{tblr}
\caption[Correlation between Soft-Score and averaged human coverage scores.]{Pearson correlation between the Soft-Score (incl. different similarity functions) and averaged human coverage scores, along with the evaluation runtime, on ArgKP21. 
For the scenario within topics and stances,  
standard deviations are indicated below the correlation values.}
\label{tab:soft_score_corr_cov}
\end{table*}

\begin{table*}[!h]
\small
\centering
\resizebox{1\textwidth}{!}{
\begin{tblr}{
  colspec = {X[c,m,wd=22mm]||X[c,m]|X[c,m,wd=22mm]||X[c,m]|X[c,m,wd=22mm]||X[c,m]|X[c,m,wd=22mm]},
  %hline{1} = {2-5}{},
  width=1\textwidth,
  stretch = 0,
  rowsep = 4pt,
  %hlines = {black, 0.5pt},
  %vlines = {black, 0.5pt}
}
\hline
\hline
\SetCell[r=2]{c} Threshold & \SetCell[c=2]{c} \textbf{CS} {(BarH)}& 
& \SetCell[c=2]{c} \textbf{CS} {(SMatchToPr)} & 
& \SetCell[c=2]{c} \textbf{CS} {(MCArgSum)} & \\
& Across & Within & Across & Within & Across & Within & \\
\hline		
0.40 & 0.478 & {0.585 {$\pm$0.254}} &
-0.092 & {-0.157 {$\pm$0.037}} & 0.206 & {-0.057 {$\pm$0.223}} \\
\hline[dotted]
0.45 & 0.475 & {0.605 {$\pm$0.248}} & 0.163 & {-0.074 {$\pm$0.187}} & 0.300 & {-0.019 {$\pm$0.307}} \\
\hline[dotted]
0.50 & 0.465 & {0.627 {$\pm$0.251}} & 0.174 & {-0.023 {$\pm$0.196}} & 0.300 & {-0.019 {$\pm$0.307}} \\
\hline[dotted]
0.55 & 0.462 & {0.657 {$\pm$0.273}} & 0.378 & {0.249 {$\pm$0.307}} & 0.281 & {-0.002 {$\pm$0.292}} \\
\hline[dotted]
0.60 & \textbf{0.489} & {\textbf{0.698} {$\pm$0.222}} & 0.469 & {0.338 {$\pm$0.371}} & 0.415 & {0.137 {$\pm$0.390}} \\
\hline[dotted]
0.65 & 0.464 & {0.676 {$\pm$0.218}} & 0.465 & {0.411 {$\pm$0.256}} & \textbf{0.449} & {0.256 {$\pm$0.357}} \\
\hline[dotted]
0.70 & 0.458 & {0.657 {$\pm$0.233}} & 0.457 & {0.404 {$\pm$0.212}} & 0.369 & {0.297 {$\pm$0.295}} \\
\hline[dotted]
0.75 & 0.466 & {0.658 {$\pm$0.197}} & \textbf{0.541} & {0.550 {$\pm$0.182}} & 0.379 & {0.347 {$\pm$0.319}} \\
\hline[dotted]
0.80 & 0.429 & {0.591 {$\pm$0.154}} & 0.511 & {\textbf{0.600} {$\pm$0.196}} & 0.444 & {\textbf{0.551} {$\pm$0.193}} \\
\hline[dotted]
0.85 & 0.414 & {0.556 {$\pm$0.201}} & 0.468 & {0.558 {$\pm$0.085}} & 0.364 & {0.421 {$\pm$0.270}} \\
\hline[dotted]
0.90 & 0.295 & {0.504 {$\pm$0.249}} & 0.238 & {0.261 {$\pm$0.070}} & 0.316 & {0.401 {$\pm$0.129}} \\
\hline
{Average\\Runtime (s)} & \SetCell[c=2]{c} 70.302 &
& \SetCell[c=2]{c} 20.961 & 
& \SetCell[c=2]{c} 14.689 & 2-6\\
\hline
\hline
\end{tblr}
}
\caption[Correlation between CS and averaged human coverage scores.]{Pearson correlation coefficient between the CS (incl. different Match Scorers) and averaged human coverage scores for different matching thresholds, along with the evaluation runtime, on ArgKP21. For the scenario within topics and stances, standard deviations are indicated alongside the correlation values.} 
\label{tab:coverage_score_corr}
\end{table*}

\input{results_tables/summarization_capability_table}

\begin{table*}[!ht]
\small
\centering
\resizebox{\linewidth}{!}{
\noindent\begin{tabularx}{\textwidth}{ >{\hsize=.2\hsize}X | >{\hsize=.4\hsize}X | >{\hsize=.4\hsize}X }
\toprule
\textbf{LLaMA-3.3-70B} & \textbf{Qwen-3-32B}                            &  \textbf{GPT-4o}                \\ \midrule
Restricts free speech      & Freedom of speech must be protected    & 
Regulation undermines freedom of expression.
\\ \midrule
Violates privacy         & Government shouldn't control private communication   & People may feel watched and censored.                                \\ \midrule
Infringes rights    & Government could abuse political power  & Government could exploit for political gain.
                 \\    \bottomrule        
\end{tabularx}}
\caption{Selected examples of generated summaries taking an against stance on the topic `Social media platforms should be regulated by the government'.}\label{tab:ex}
\end{table*}

%% file: results_tables/summarization_capability_table.tex
\begin{table*}[!ht]
\small
\resizebox{\textwidth}{!}{%
\begin{tabular}{@{}lcccccc@{}}
\toprule
                          & \multicolumn{3}{c|}{ArgKP21}                          & \multicolumn{3}{c}{Debate}       \\ \midrule
                    & Coverage & Redundancy & \multicolumn{1}{l|}{Weighted} & Coverage & Redundancy & Weighted \\ \midrule
\multicolumn{7}{l}{\emph{Classification-based}   }                                                                          \\ \midrule
BarH                      & 0.819    & 0.093      & \multicolumn{1}{l|}{0.848}    & 0.764    & 0.218      & 0.770    \\
BarH+cand(gpt-4o)         & 0.904    & 0.177      & \multicolumn{1}{l|}{0.877}    & 0.843    & 0.146      & 0.847    \\
BarH+cand(llama-3.3-70b)  & 0.829    & 0.125      & \multicolumn{1}{l|}{0.845}    & 0.806    & 0.192      & 0.807    \\
BarH+cand(qwen-2.5-72b)   & 0.830    & 0.173      & \multicolumn{1}{l|}{0.829}    & 0.825    & 0.228      & 0.807    \\
BarH+cand(qwen3-32b)      & 0.915    & 0.129      & \multicolumn{1}{l|}{0.900}    & 0.891    & 0.142      & 0.880    \\ \midrule
SMtPR                     & 0.905    & 0.240      & \multicolumn{1}{l|}{0.856}    & 0.780    & 0.147      & 0.805    \\
SMtPR+cand(gpt-4o)        & 0.912    & 0.172      & \multicolumn{1}{l|}{0.884}    & 0.862    & 0.116      & 0.869    \\
SMtPR+cand(llama-3.3-70b) & 0.898    & 0.348      & \multicolumn{1}{l|}{0.816}    & 0.893    & 0.343      & 0.815    \\
SMtPR+cand(qwen-2.5-72b)  & 0.853    & 0.176      & \multicolumn{1}{l|}{0.843}    & 0.864    & 0.150      & 0.860    \\
SMtPR+cand(qwen3-32b)     & 0.933    & 0.177      & \multicolumn{1}{l|}{0.896}    & 0.922    & 0.173      & 0.890    \\ \midrule
\multicolumn{7}{l}{\emph{Clustering-based}    }                                                                             \\ \midrule
USKPM                     & 0.824    & 0.249      & \multicolumn{1}{l|}{0.800}    & 0.806    & 0.112      & 0.833    \\
MCArgSum(gpt-4o)          & 0.844    & 0.156      & \multicolumn{1}{l|}{0.844}    & 0.884    & 0.112      & 0.886    \\
MCArgSum(llama-3.3-70b)   & 0.713    & 0.132      & \multicolumn{1}{l|}{0.765}    & 0.636    & 0.084      & 0.729    \\
MCArgSum(qwen-2.5-72b)    & 0.809    & 0.079      & \multicolumn{1}{l|}{0.847}    & 0.887    & 0.134      & 0.880    \\
MCArgSum(qwen3-32b)       & 0.839    & 0.119      & \multicolumn{1}{l|}{0.853}    & 0.896    & 0.099      & 0.898    \\ \midrule
\multicolumn{7}{l}{\emph{LLM end-to-end Summarization}    }   
                                                        \\ \midrule
gpt-4o          & 0.808 & 0.048      & \multicolumn{1}{l|}{0.856}    & 0.791 & 0.060 & 0.841    \\
llama-3.3-70b   & 0.840 & 0.269      & \multicolumn{1}{l|}{0.804}    & 0.835 & 0.301 & 0.790    \\
qwen-2.5-72b    & 0.900 & 0.086      & \multicolumn{1}{l|}{0.904}    & 0.850 & 0.126 & 0.858    \\
qwen3-32b       & 0.934 & 0.099      & \multicolumn{1}{l|}{0.923}    
 & 0.892 & 0.089 & 0.899    \\ \bottomrule
\end{tabular}}
\caption{Coverage and Redundancy Scores as well as Weighted Scores for ArgKP21 and Debate datasets. For BarH and SMatchToPr (abbreviated as SMtPR), the variant with LLM-based candidates is indicated by +cand and the models in brackets indicate the LLMs integrated.}\label{tab:sum}
\end{table*}%\todo{YC: bold and underline some results}

%% file: acl2023.bbl
\begin{thebibliography}{54}
\expandafter\ifx\csname natexlab\endcsname\relax\def\natexlab#1{#1}\fi

\bibitem[{Ajjour et~al.(2019)Ajjour, Alshomary, Wachsmuth, and Stein}]{ajjour-etal-2019-modeling}
Yamen Ajjour, Milad Alshomary, Henning Wachsmuth, and Benno Stein. 2019.
\newblock \href {https://doi.org/10.18653/v1/D19-1290} {Modeling frames in argumentation}.
\newblock In \emph{Proceedings of the 2019 Conference on Empirical Methods in Natural Language Processing and the 9th International Joint Conference on Natural Language Processing (EMNLP-IJCNLP)}, pages 2922--2932, Hong Kong, China. Association for Computational Linguistics.

\bibitem[{Alshomary et~al.(2021)Alshomary, Gurcke, Syed, Heinisch, Splieth{\"o}ver, Cimiano, Potthast, and Wachsmuth}]{alshomary-etal-2021-key}
Milad Alshomary, Timon Gurcke, Shahbaz Syed, Philipp Heinisch, Maximilian Splieth{\"o}ver, Philipp Cimiano, Martin Potthast, and Henning Wachsmuth. 2021.
\newblock \href {https://doi.org/10.18653/v1/2021.argmining-1.19} {Key point analysis via contrastive learning and extractive argument summarization}.
\newblock In \emph{Proceedings of the 8th Workshop on Argument Mining}, pages 184--189, Punta Cana, Dominican Republic. Association for Computational Linguistics.

\bibitem[{Anand and Wagh(2022)}]{ANAND20222141}
Deepa Anand and Rupali Wagh. 2022.
\newblock \href {https://doi.org/https://doi.org/10.1016/j.jksuci.2019.11.015} {Effective deep learning approaches for summarization of legal texts}.
\newblock \emph{Journal of King Saud University - Computer and Information Sciences}, 34(5):2141--2150.

\bibitem[{Bar-Haim et~al.(2020{\natexlab{a}})Bar-Haim, Eden, Friedman, Kantor, Lahav, and Slonim}]{bar-haim-etal-2020-arguments}
Roy Bar-Haim, Lilach Eden, Roni Friedman, Yoav Kantor, Dan Lahav, and Noam Slonim. 2020{\natexlab{a}}.
\newblock \href {https://doi.org/10.18653/v1/2020.acl-main.371} {From arguments to key points: {T}owards automatic argument summarization}.
\newblock In \emph{Proceedings of the 58th Annual Meeting of the Association for Computational Linguistics}, pages 4029--4039, Online. Association for Computational Linguistics.

\bibitem[{Bar-Haim et~al.(2020{\natexlab{b}})Bar-Haim, Kantor, Eden, Friedman, Lahav, and Slonim}]{bar-haim-etal-2020-quantitative}
Roy Bar-Haim, Yoav Kantor, Lilach Eden, Roni Friedman, Dan Lahav, and Noam Slonim. 2020{\natexlab{b}}.
\newblock \href {https://doi.org/10.18653/v1/2020.emnlp-main.3} {Quantitative argument summarization and beyond: Cross-domain key point analysis}.
\newblock In \emph{Proceedings of the 2020 Conference on Empirical Methods in Natural Language Processing (EMNLP)}, pages 39--49, Online. Association for Computational Linguistics.

\bibitem[{Celikyilmaz et~al.(2021)Celikyilmaz, Clark, and Gao}]{celikyilmaz2021}
Asli Celikyilmaz, Elizabeth Clark, and Jianfeng Gao. 2021.
\newblock \href {http://arxiv.org/abs/2006.14799} {Evaluation of text generation: A survey}.

\bibitem[{Chang et~al.(2024)Chang, Wang, Wang, Wu, Yang, Zhu, Chen, Yi, Wang, Wang, Ye, Zhang, Chang, Yu, Yang, and Xie}]{10.1145/3641289}
Yupeng Chang, Xu~Wang, Jindong Wang, Yuan Wu, Linyi Yang, Kaijie Zhu, Hao Chen, Xiaoyuan Yi, Cunxiang Wang, Yidong Wang, Wei Ye, Yue Zhang, Yi~Chang, Philip~S. Yu, Qiang Yang, and Xing Xie. 2024.
\newblock \href {https://doi.org/10.1145/3641289} {A survey on evaluation of large language models}.
\newblock \emph{ACM Trans. Intell. Syst. Technol.}, 15(3).

\bibitem[{Chen and Eger(2023)}]{chen-eger-2023-menli}
Yanran Chen and Steffen Eger. 2023.
\newblock \href {https://doi.org/10.1162/tacl_a_00576} {{MENLI}: Robust evaluation metrics from natural language inference}.
\newblock \emph{Transactions of the Association for Computational Linguistics}, 11:804--825.

\bibitem[{Chung et~al.(2022)Chung, Hou, Longpre, Zoph, Tay, Fedus, Li, Wang, Dehghani, Brahma, Webson, Gu, Dai, Suzgun, Chen, Chowdhery, Narang, Mishra, Yu, Zhao, Huang, Dai, Yu, Petrov, Chi, Dean, Devlin, Roberts, Zhou, Le, and Wei}]{flant5}
Hyung~Won Chung, Le~Hou, Shayne Longpre, Barret Zoph, Yi~Tay, William Fedus, Eric Li, Xuezhi Wang, Mostafa Dehghani, Siddhartha Brahma, Albert Webson, Shixiang~Shane Gu, Zhuyun Dai, Mirac Suzgun, Xinyun Chen, Aakanksha Chowdhery, Sharan Narang, Gaurav Mishra, Adams Yu, Vincent Zhao, Yanping Huang, Andrew Dai, Hongkun Yu, Slav Petrov, Ed~H. Chi, Jeff Dean, Jacob Devlin, Adam Roberts, Denny Zhou, Quoc~V. Le, and Jason Wei. 2022.
\newblock \href {https://doi.org/10.48550/ARXIV.2210.11416} {Scaling instruction-finetuned language models}.

\bibitem[{Day and Edelsbrunner(1984)}]{Day1984EfficientAF}
William H.~E. Day and Herbert Edelsbrunner. 1984.
\newblock \href {https://api.semanticscholar.org/CorpusID:121201396} {Efficient algorithms for agglomerative hierarchical clustering methods}.
\newblock \emph{Journal of Classification}, 1:7--24.

\bibitem[{Es et~al.(2024)Es, James, Espinosa~Anke, and Schockaert}]{es-etal-2024-ragas}
Shahul Es, Jithin James, Luis Espinosa~Anke, and Steven Schockaert. 2024.
\newblock \href {https://aclanthology.org/2024.eacl-demo.16/} {{RAGA}s: Automated evaluation of retrieval augmented generation}.
\newblock In \emph{Proceedings of the 18th Conference of the European Chapter of the Association for Computational Linguistics: System Demonstrations}, pages 150--158, St. Julians, Malta. Association for Computational Linguistics.

\bibitem[{Fernandes et~al.(2023)Fernandes, Deutsch, Finkelstein, Riley, Martins, Neubig, Garg, Clark, Freitag, and Firat}]{fernandes-etal-2023-devil}
Patrick Fernandes, Daniel Deutsch, Mara Finkelstein, Parker Riley, Andr{\'e} Martins, Graham Neubig, Ankush Garg, Jonathan Clark, Markus Freitag, and Orhan Firat. 2023.
\newblock \href {https://doi.org/10.18653/v1/2023.wmt-1.100} {The devil is in the errors: Leveraging large language models for fine-grained machine translation evaluation}.
\newblock In \emph{Proceedings of the Eighth Conference on Machine Translation}, pages 1066--1083, Singapore. Association for Computational Linguistics.

\bibitem[{Friedman et~al.(2021)Friedman, Dankin, Hou, Aharonov, Katz, and Slonim}]{friedman-etal-2021-overview}
Roni Friedman, Lena Dankin, Yufang Hou, Ranit Aharonov, Yoav Katz, and Noam Slonim. 2021.
\newblock \href {https://doi.org/10.18653/v1/2021.argmining-1.16} {Overview of the 2021 key point analysis shared task}.
\newblock In \emph{Proceedings of the 8th Workshop on Argument Mining}, pages 154--164, Punta Cana, Dominican Republic. Association for Computational Linguistics.

\bibitem[{Fu et~al.(2024)Fu, Ng, Jiang, and Liu}]{fu-etal-2024-gptscore}
Jinlan Fu, See-Kiong Ng, Zhengbao Jiang, and Pengfei Liu. 2024.
\newblock \href {https://doi.org/10.18653/v1/2024.naacl-long.365} {{GPTS}core: Evaluate as you desire}.
\newblock In \emph{Proceedings of the 2024 Conference of the North American Chapter of the Association for Computational Linguistics: Human Language Technologies (Volume 1: Long Papers)}, pages 6556--6576, Mexico City, Mexico. Association for Computational Linguistics.

\bibitem[{Giarelis et~al.(2023)Giarelis, Mastrokostas, and Karacapilidis}]{app13137620}
Nikolaos Giarelis, Charalampos Mastrokostas, and Nikos Karacapilidis. 2023.
\newblock \href {https://doi.org/10.3390/app13137620} {Abstractive vs. extractive summarization: An experimental review}.
\newblock \emph{Applied Sciences}, 13(13).

\bibitem[{Grattafiori et~al.(2024)Grattafiori, Dubey, Jauhri, Pandey, Kadian, Al-Dahle, Letman, Mathur, Schelten, Vaughan, Yang, Fan, Goyal, Hartshorn, Yang, Mitra, Sravankumar, Korenev, Hinsvark, Rao, Zhang, Rodriguez, Gregerson, Spataru, Roziere, Biron, Tang, Chern, Caucheteux, Nayak, Bi, Marra, McConnell, Keller, Touret, Wu, Wong, Ferrer, Nikolaidis, Allonsius, Song, Pintz, Livshits, Wyatt, Esiobu, Choudhary, Mahajan, Garcia-Olano, Perino, Hupkes, Lakomkin, AlBadawy, Lobanova, Dinan, Smith, Radenovic, Guzmán, Zhang, Synnaeve, Lee, Anderson, Thattai, Nail, Mialon, Pang, Cucurell, Nguyen, Korevaar, Xu, Touvron, Zarov, Ibarra, Kloumann, Misra, Evtimov, Zhang, Copet, Lee, Geffert, Vranes, Park, Mahadeokar, Shah, van~der Linde, Billock, Hong, Lee, Fu, Chi, Huang, Liu, Wang, Yu, Bitton, Spisak, Park, Rocca, Johnstun, Saxe, Jia, Alwala, Prasad, Upasani, Plawiak, Li, Heafield, Stone, El-Arini, Iyer, Malik, Chiu, Bhalla, Lakhotia, Rantala-Yeary, van~der Maaten, Chen, Tan, Jenkins, Martin, Madaan, Malo, Blecher,
  Landzaat, de~Oliveira, Muzzi, Pasupuleti, Singh, Paluri, Kardas, Tsimpoukelli, Oldham, Rita, Pavlova, Kambadur, Lewis, Si, Singh, Hassan, Goyal, Torabi, Bashlykov, Bogoychev, Chatterji, Zhang, Duchenne, Çelebi, Alrassy, Zhang, Li, Vasic, Weng, Bhargava, Dubal, Krishnan, Koura, Xu, He, Dong, Srinivasan, Ganapathy, Calderer, Cabral, Stojnic, Raileanu, Maheswari, Girdhar, Patel, Sauvestre, Polidoro, Sumbaly, Taylor, Silva, Hou, Wang, Hosseini, Chennabasappa, Singh, Bell, Kim, Edunov, Nie, Narang, Raparthy, Shen, Wan, Bhosale, Zhang, Vandenhende, Batra, Whitman, Sootla, Collot, Gururangan, Borodinsky, Herman, Fowler, Sheasha, Georgiou, Scialom, Speckbacher, Mihaylov, Xiao, Karn, Goswami, Gupta, Ramanathan, Kerkez, Gonguet, Do, Vogeti, Albiero, Petrovic, Chu, Xiong, Fu, Meers, Martinet, Wang, Wang, Tan, Xia, Xie, Jia, Wang, Goldschlag, Gaur, Babaei, Wen, Song, Zhang, Li, Mao, Coudert, Yan, Chen, Papakipos, Singh, Srivastava, Jain, Kelsey, Shajnfeld, Gangidi, Victoria, Goldstand, Menon, Sharma, Boesenberg,
  Baevski, Feinstein, Kallet, Sangani, Teo, Yunus, Lupu, Alvarado, Caples, Gu, Ho, Poulton, Ryan, Ramchandani, Dong, Franco, Goyal, Saraf, Chowdhury, Gabriel, Bharambe, Eisenman, Yazdan, James, Maurer, Leonhardi, Huang, Loyd, Paola, Paranjape, Liu, Wu, Ni, Hancock, Wasti, Spence, Stojkovic, Gamido, Montalvo, Parker, Burton, Mejia, Liu, Wang, Kim, Zhou, Hu, Chu, Cai, Tindal, Feichtenhofer, Gao, Civin, Beaty, Kreymer, Li, Adkins, Xu, Testuggine, David, Parikh, Liskovich, Foss, Wang, Le, Holland, Dowling, Jamil, Montgomery, Presani, Hahn, Wood, Le, Brinkman, Arcaute, Dunbar, Smothers, Sun, Kreuk, Tian, Kokkinos, Ozgenel, Caggioni, Kanayet, Seide, Florez, Schwarz, Badeer, Swee, Halpern, Herman, Sizov, Guangyi, Zhang, Lakshminarayanan, Inan, Shojanazeri, Zou, Wang, Zha, Habeeb, Rudolph, Suk, Aspegren, Goldman, Zhan, Damlaj, Molybog, Tufanov, Leontiadis, Veliche, Gat, Weissman, Geboski, Kohli, Lam, Asher, Gaya, Marcus, Tang, Chan, Zhen, Reizenstein, Teboul, Zhong, Jin, Yang, Cummings, Carvill, Shepard, McPhie,
  Torres, Ginsburg, Wang, Wu, U, Saxena, Khandelwal, Zand, Matosich, Veeraraghavan, Michelena, Li, Jagadeesh, Huang, Chawla, Huang, Chen, Garg, A, Silva, Bell, Zhang, Guo, Yu, Moshkovich, Wehrstedt, Khabsa, Avalani, Bhatt, Mankus, Hasson, Lennie, Reso, Groshev, Naumov, Lathi, Keneally, Liu, Seltzer, Valko, Restrepo, Patel, Vyatskov, Samvelyan, Clark, Macey, Wang, Hermoso, Metanat, Rastegari, Bansal, Santhanam, Parks, White, Bawa, Singhal, Egebo, Usunier, Mehta, Laptev, Dong, Cheng, Chernoguz, Hart, Salpekar, Kalinli, Kent, Parekh, Saab, Balaji, Rittner, Bontrager, Roux, Dollar, Zvyagina, Ratanchandani, Yuvraj, Liang, Alao, Rodriguez, Ayub, Murthy, Nayani, Mitra, Parthasarathy, Li, Hogan, Battey, Wang, Howes, Rinott, Mehta, Siby, Bondu, Datta, Chugh, Hunt, Dhillon, Sidorov, Pan, Mahajan, Verma, Yamamoto, Ramaswamy, Lindsay, Lindsay, Feng, Lin, Zha, Patil, Shankar, Zhang, Zhang, Wang, Agarwal, Sajuyigbe, Chintala, Max, Chen, Kehoe, Satterfield, Govindaprasad, Gupta, Deng, Cho, Virk, Subramanian, Choudhury,
  Goldman, Remez, Glaser, Best, Koehler, Robinson, Li, Zhang, Matthews, Chou, Shaked, Vontimitta, Ajayi, Montanez, Mohan, Kumar, Mangla, Ionescu, Poenaru, Mihailescu, Ivanov, Li, Wang, Jiang, Bouaziz, Constable, Tang, Wu, Wang, Wu, Gao, Kleinman, Chen, Hu, Jia, Qi, Li, Zhang, Zhang, Adi, Nam, Yu, Wang, Zhao, Hao, Qian, Li, He, Rait, DeVito, Rosnbrick, Wen, Yang, Zhao, and Ma}]{grattafiori2024llama3herdmodels}
Aaron Grattafiori, Abhimanyu Dubey, Abhinav Jauhri, Abhinav Pandey, Abhishek Kadian, Ahmad Al-Dahle, Aiesha Letman, Akhil Mathur, Alan Schelten, Alex Vaughan, Amy Yang, Angela Fan, Anirudh Goyal, Anthony Hartshorn, Aobo Yang, Archi Mitra, Archie Sravankumar, Artem Korenev, Arthur Hinsvark, Arun Rao, Aston Zhang, Aurelien Rodriguez, Austen Gregerson, Ava Spataru, Baptiste Roziere, Bethany Biron, Binh Tang, Bobbie Chern, Charlotte Caucheteux, Chaya Nayak, Chloe Bi, Chris Marra, Chris McConnell, Christian Keller, Christophe Touret, Chunyang Wu, Corinne Wong, Cristian~Canton Ferrer, Cyrus Nikolaidis, Damien Allonsius, Daniel Song, Danielle Pintz, Danny Livshits, Danny Wyatt, David Esiobu, Dhruv Choudhary, Dhruv Mahajan, Diego Garcia-Olano, Diego Perino, Dieuwke Hupkes, Egor Lakomkin, Ehab AlBadawy, Elina Lobanova, Emily Dinan, Eric~Michael Smith, Filip Radenovic, Francisco Guzmán, Frank Zhang, Gabriel Synnaeve, Gabrielle Lee, Georgia~Lewis Anderson, Govind Thattai, Graeme Nail, Gregoire Mialon, Guan Pang,
  Guillem Cucurell, Hailey Nguyen, Hannah Korevaar, Hu~Xu, Hugo Touvron, Iliyan Zarov, Imanol~Arrieta Ibarra, Isabel Kloumann, Ishan Misra, Ivan Evtimov, Jack Zhang, Jade Copet, Jaewon Lee, Jan Geffert, Jana Vranes, Jason Park, Jay Mahadeokar, Jeet Shah, Jelmer van~der Linde, Jennifer Billock, Jenny Hong, Jenya Lee, Jeremy Fu, Jianfeng Chi, Jianyu Huang, Jiawen Liu, Jie Wang, Jiecao Yu, Joanna Bitton, Joe Spisak, Jongsoo Park, Joseph Rocca, Joshua Johnstun, Joshua Saxe, Junteng Jia, Kalyan~Vasuden Alwala, Karthik Prasad, Kartikeya Upasani, Kate Plawiak, Ke~Li, Kenneth Heafield, Kevin Stone, Khalid El-Arini, Krithika Iyer, Kshitiz Malik, Kuenley Chiu, Kunal Bhalla, Kushal Lakhotia, Lauren Rantala-Yeary, Laurens van~der Maaten, Lawrence Chen, Liang Tan, Liz Jenkins, Louis Martin, Lovish Madaan, Lubo Malo, Lukas Blecher, Lukas Landzaat, Luke de~Oliveira, Madeline Muzzi, Mahesh Pasupuleti, Mannat Singh, Manohar Paluri, Marcin Kardas, Maria Tsimpoukelli, Mathew Oldham, Mathieu Rita, Maya Pavlova, Melanie Kambadur,
  Mike Lewis, Min Si, Mitesh~Kumar Singh, Mona Hassan, Naman Goyal, Narjes Torabi, Nikolay Bashlykov, Nikolay Bogoychev, Niladri Chatterji, Ning Zhang, Olivier Duchenne, Onur Çelebi, Patrick Alrassy, Pengchuan Zhang, Pengwei Li, Petar Vasic, Peter Weng, Prajjwal Bhargava, Pratik Dubal, Praveen Krishnan, Punit~Singh Koura, Puxin Xu, Qing He, Qingxiao Dong, Ragavan Srinivasan, Raj Ganapathy, Ramon Calderer, Ricardo~Silveira Cabral, Robert Stojnic, Roberta Raileanu, Rohan Maheswari, Rohit Girdhar, Rohit Patel, Romain Sauvestre, Ronnie Polidoro, Roshan Sumbaly, Ross Taylor, Ruan Silva, Rui Hou, Rui Wang, Saghar Hosseini, Sahana Chennabasappa, Sanjay Singh, Sean Bell, Seohyun~Sonia Kim, Sergey Edunov, Shaoliang Nie, Sharan Narang, Sharath Raparthy, Sheng Shen, Shengye Wan, Shruti Bhosale, Shun Zhang, Simon Vandenhende, Soumya Batra, Spencer Whitman, Sten Sootla, Stephane Collot, Suchin Gururangan, Sydney Borodinsky, Tamar Herman, Tara Fowler, Tarek Sheasha, Thomas Georgiou, Thomas Scialom, Tobias Speckbacher,
  Todor Mihaylov, Tong Xiao, Ujjwal Karn, Vedanuj Goswami, Vibhor Gupta, Vignesh Ramanathan, Viktor Kerkez, Vincent Gonguet, Virginie Do, Vish Vogeti, Vítor Albiero, Vladan Petrovic, Weiwei Chu, Wenhan Xiong, Wenyin Fu, Whitney Meers, Xavier Martinet, Xiaodong Wang, Xiaofang Wang, Xiaoqing~Ellen Tan, Xide Xia, Xinfeng Xie, Xuchao Jia, Xuewei Wang, Yaelle Goldschlag, Yashesh Gaur, Yasmine Babaei, Yi~Wen, Yiwen Song, Yuchen Zhang, Yue Li, Yuning Mao, Zacharie~Delpierre Coudert, Zheng Yan, Zhengxing Chen, Zoe Papakipos, Aaditya Singh, Aayushi Srivastava, Abha Jain, Adam Kelsey, Adam Shajnfeld, Adithya Gangidi, Adolfo Victoria, Ahuva Goldstand, Ajay Menon, Ajay Sharma, Alex Boesenberg, Alexei Baevski, Allie Feinstein, Amanda Kallet, Amit Sangani, Amos Teo, Anam Yunus, Andrei Lupu, Andres Alvarado, Andrew Caples, Andrew Gu, Andrew Ho, Andrew Poulton, Andrew Ryan, Ankit Ramchandani, Annie Dong, Annie Franco, Anuj Goyal, Aparajita Saraf, Arkabandhu Chowdhury, Ashley Gabriel, Ashwin Bharambe, Assaf Eisenman, Azadeh
  Yazdan, Beau James, Ben Maurer, Benjamin Leonhardi, Bernie Huang, Beth Loyd, Beto~De Paola, Bhargavi Paranjape, Bing Liu, Bo~Wu, Boyu Ni, Braden Hancock, Bram Wasti, Brandon Spence, Brani Stojkovic, Brian Gamido, Britt Montalvo, Carl Parker, Carly Burton, Catalina Mejia, Ce~Liu, Changhan Wang, Changkyu Kim, Chao Zhou, Chester Hu, Ching-Hsiang Chu, Chris Cai, Chris Tindal, Christoph Feichtenhofer, Cynthia Gao, Damon Civin, Dana Beaty, Daniel Kreymer, Daniel Li, David Adkins, David Xu, Davide Testuggine, Delia David, Devi Parikh, Diana Liskovich, Didem Foss, Dingkang Wang, Duc Le, Dustin Holland, Edward Dowling, Eissa Jamil, Elaine Montgomery, Eleonora Presani, Emily Hahn, Emily Wood, Eric-Tuan Le, Erik Brinkman, Esteban Arcaute, Evan Dunbar, Evan Smothers, Fei Sun, Felix Kreuk, Feng Tian, Filippos Kokkinos, Firat Ozgenel, Francesco Caggioni, Frank Kanayet, Frank Seide, Gabriela~Medina Florez, Gabriella Schwarz, Gada Badeer, Georgia Swee, Gil Halpern, Grant Herman, Grigory Sizov, Guangyi, Zhang, Guna
  Lakshminarayanan, Hakan Inan, Hamid Shojanazeri, Han Zou, Hannah Wang, Hanwen Zha, Haroun Habeeb, Harrison Rudolph, Helen Suk, Henry Aspegren, Hunter Goldman, Hongyuan Zhan, Ibrahim Damlaj, Igor Molybog, Igor Tufanov, Ilias Leontiadis, Irina-Elena Veliche, Itai Gat, Jake Weissman, James Geboski, James Kohli, Janice Lam, Japhet Asher, Jean-Baptiste Gaya, Jeff Marcus, Jeff Tang, Jennifer Chan, Jenny Zhen, Jeremy Reizenstein, Jeremy Teboul, Jessica Zhong, Jian Jin, Jingyi Yang, Joe Cummings, Jon Carvill, Jon Shepard, Jonathan McPhie, Jonathan Torres, Josh Ginsburg, Junjie Wang, Kai Wu, Kam~Hou U, Karan Saxena, Kartikay Khandelwal, Katayoun Zand, Kathy Matosich, Kaushik Veeraraghavan, Kelly Michelena, Keqian Li, Kiran Jagadeesh, Kun Huang, Kunal Chawla, Kyle Huang, Lailin Chen, Lakshya Garg, Lavender A, Leandro Silva, Lee Bell, Lei Zhang, Liangpeng Guo, Licheng Yu, Liron Moshkovich, Luca Wehrstedt, Madian Khabsa, Manav Avalani, Manish Bhatt, Martynas Mankus, Matan Hasson, Matthew Lennie, Matthias Reso, Maxim
  Groshev, Maxim Naumov, Maya Lathi, Meghan Keneally, Miao Liu, Michael~L. Seltzer, Michal Valko, Michelle Restrepo, Mihir Patel, Mik Vyatskov, Mikayel Samvelyan, Mike Clark, Mike Macey, Mike Wang, Miquel~Jubert Hermoso, Mo~Metanat, Mohammad Rastegari, Munish Bansal, Nandhini Santhanam, Natascha Parks, Natasha White, Navyata Bawa, Nayan Singhal, Nick Egebo, Nicolas Usunier, Nikhil Mehta, Nikolay~Pavlovich Laptev, Ning Dong, Norman Cheng, Oleg Chernoguz, Olivia Hart, Omkar Salpekar, Ozlem Kalinli, Parkin Kent, Parth Parekh, Paul Saab, Pavan Balaji, Pedro Rittner, Philip Bontrager, Pierre Roux, Piotr Dollar, Polina Zvyagina, Prashant Ratanchandani, Pritish Yuvraj, Qian Liang, Rachad Alao, Rachel Rodriguez, Rafi Ayub, Raghotham Murthy, Raghu Nayani, Rahul Mitra, Rangaprabhu Parthasarathy, Raymond Li, Rebekkah Hogan, Robin Battey, Rocky Wang, Russ Howes, Ruty Rinott, Sachin Mehta, Sachin Siby, Sai~Jayesh Bondu, Samyak Datta, Sara Chugh, Sara Hunt, Sargun Dhillon, Sasha Sidorov, Satadru Pan, Saurabh Mahajan,
  Saurabh Verma, Seiji Yamamoto, Sharadh Ramaswamy, Shaun Lindsay, Shaun Lindsay, Sheng Feng, Shenghao Lin, Shengxin~Cindy Zha, Shishir Patil, Shiva Shankar, Shuqiang Zhang, Shuqiang Zhang, Sinong Wang, Sneha Agarwal, Soji Sajuyigbe, Soumith Chintala, Stephanie Max, Stephen Chen, Steve Kehoe, Steve Satterfield, Sudarshan Govindaprasad, Sumit Gupta, Summer Deng, Sungmin Cho, Sunny Virk, Suraj Subramanian, Sy~Choudhury, Sydney Goldman, Tal Remez, Tamar Glaser, Tamara Best, Thilo Koehler, Thomas Robinson, Tianhe Li, Tianjun Zhang, Tim Matthews, Timothy Chou, Tzook Shaked, Varun Vontimitta, Victoria Ajayi, Victoria Montanez, Vijai Mohan, Vinay~Satish Kumar, Vishal Mangla, Vlad Ionescu, Vlad Poenaru, Vlad~Tiberiu Mihailescu, Vladimir Ivanov, Wei Li, Wenchen Wang, Wenwen Jiang, Wes Bouaziz, Will Constable, Xiaocheng Tang, Xiaojian Wu, Xiaolan Wang, Xilun Wu, Xinbo Gao, Yaniv Kleinman, Yanjun Chen, Ye~Hu, Ye~Jia, Ye~Qi, Yenda Li, Yilin Zhang, Ying Zhang, Yossi Adi, Youngjin Nam, Yu, Wang, Yu~Zhao, Yuchen Hao, Yundi
  Qian, Yunlu Li, Yuzi He, Zach Rait, Zachary DeVito, Zef Rosnbrick, Zhaoduo Wen, Zhenyu Yang, Zhiwei Zhao, and Zhiyu Ma. 2024.
\newblock \href {http://arxiv.org/abs/2407.21783} {The llama 3 herd of models}.

\bibitem[{Gretz et~al.(2020)Gretz, Friedman, Cohen{-}Karlik, Toledo, Lahav, Aharonov, and Slonim}]{DBLP:conf/aaai/GretzFCTLAS20}
Shai Gretz, Roni Friedman, Edo Cohen{-}Karlik, Assaf Toledo, Dan Lahav, Ranit Aharonov, and Noam Slonim. 2020.
\newblock \href {https://doi.org/10.1609/AAAI.V34I05.6285} {A large-scale dataset for argument quality ranking: Construction and analysis}.
\newblock In \emph{The Thirty-Fourth {AAAI} Conference on Artificial Intelligence, {AAAI} 2020, The Thirty-Second Innovative Applications of Artificial Intelligence Conference, {IAAI} 2020, The Tenth {AAAI} Symposium on Educational Advances in Artificial Intelligence, {EAAI} 2020, New York, NY, USA, February 7-12, 2020}, pages 7805--7813. {AAAI} Press.

\bibitem[{Grootendorst(2022)}]{grootendorst2022bertopicneuraltopicmodeling}
Maarten Grootendorst. 2022.
\newblock \href {http://arxiv.org/abs/2203.05794} {Bertopic: Neural topic modeling with a class-based tf-idf procedure}.

\bibitem[{Hasan and Ng(2014)}]{debate_dataset}
Kazi~Saidul Hasan and Vincent Ng. 2014.
\newblock \href {https://doi.org/10.3115/v1/D14-1083} {Why are you taking this stance? identifying and classifying reasons in ideological debates}.
\newblock In \emph{Proceedings of the 2014 Conference on Empirical Methods in Natural Language Processing ({EMNLP})}, pages 751--762, Doha, Qatar. Association for Computational Linguistics.

\bibitem[{Khosravani et~al.(2024)Khosravani, Huang, and Trabelsi}]{khosravani-etal-2024-enhancing}
Mohammad Khosravani, Chenyang Huang, and Amine Trabelsi. 2024.
\newblock \href {https://doi.org/10.18653/v1/2024.naacl-long.454} {Enhancing argument summarization: Prioritizing exhaustiveness in key point generation and introducing an automatic coverage evaluation metric}.
\newblock In \emph{Proceedings of the 2024 Conference of the North American Chapter of the Association for Computational Linguistics: Human Language Technologies (Volume 1: Long Papers)}, pages 8212--8224, Mexico City, Mexico. Association for Computational Linguistics.

\bibitem[{Kocmi and Federmann(2023)}]{kocmi-federmann-2023-large}
Tom Kocmi and Christian Federmann. 2023.
\newblock \href {https://aclanthology.org/2023.eamt-1.19/} {Large language models are state-of-the-art evaluators of translation quality}.
\newblock In \emph{Proceedings of the 24th Annual Conference of the European Association for Machine Translation}, pages 193--203, Tampere, Finland. European Association for Machine Translation.

\bibitem[{Kusner et~al.(2015)Kusner, Sun, Kolkin, and Weinberger}]{pmlr-v37-kusnerb15}
Matt Kusner, Yu~Sun, Nicholas Kolkin, and Kilian Weinberger. 2015.
\newblock \href {https://proceedings.mlr.press/v37/kusnerb15.html} {From word embeddings to document distances}.
\newblock In \emph{Proceedings of the 32nd International Conference on Machine Learning}, volume~37 of \emph{Proceedings of Machine Learning Research}, pages 957--966, Lille, France. PMLR.

\bibitem[{Larionov and Eger(2025)}]{larionov-eger-2025-promptoptme}
Daniil Larionov and Steffen Eger. 2025.
\newblock \href {https://doi.org/10.18653/v1/2025.naacl-long.592} {{P}rompt{O}pt{M}e: Error-aware prompt compression for {LLM}-based {MT} evaluation metrics}.
\newblock In \emph{Proceedings of the 2025 Conference of the Nations of the Americas Chapter of the Association for Computational Linguistics: Human Language Technologies (Volume 1: Long Papers)}, pages 11807--11820, Albuquerque, New Mexico. Association for Computational Linguistics.

\bibitem[{Leiter and Eger(2024)}]{leiter-eger-2024-prexme}
Christoph Leiter and Steffen Eger. 2024.
\newblock \href {https://doi.org/10.18653/v1/2024.emnlp-main.641} {{P}r{E}x{M}e! large scale prompt exploration of open source {LLM}s for machine translation and summarization evaluation}.
\newblock In \emph{Proceedings of the 2024 Conference on Empirical Methods in Natural Language Processing}, pages 11481--11506, Miami, Florida, USA. Association for Computational Linguistics.

\bibitem[{Leiter et~al.(2023)Leiter, Opitz, Deutsch, Gao, Dror, and Eger}]{leiter-etal-2023-eval4nlp}
Christoph Leiter, Juri Opitz, Daniel Deutsch, Yang Gao, Rotem Dror, and Steffen Eger. 2023.
\newblock \href {https://doi.org/10.18653/v1/2023.eval4nlp-1.10} {The {E}val4{NLP} 2023 shared task on prompting large language models as explainable metrics}.
\newblock In \emph{Proceedings of the 4th Workshop on Evaluation and Comparison of NLP Systems}, pages 117--138, Bali, Indonesia. Association for Computational Linguistics.

\bibitem[{Li et~al.(2023)Li, Schlegel, Batista-Navarro, and Nenadic}]{li-etal-2023-hear}
Hao Li, Viktor Schlegel, Riza Batista-Navarro, and Goran Nenadic. 2023.
\newblock \href {https://doi.org/10.18653/v1/2023.acl-long.786} {Do you hear the people sing? key point analysis via iterative clustering and abstractive summarisation}.
\newblock In \emph{Proceedings of the 61st Annual Meeting of the Association for Computational Linguistics (Volume 1: Long Papers)}, pages 14064--14080, Toronto, Canada. Association for Computational Linguistics.

\bibitem[{Li et~al.(2024)Li, Wu, Schlegel, Batista-Navarro, Madusanka, Zahid, Zeng, Wang, He, Li, and Nenadic}]{li-etal-2024-side}
Hao Li, Yuping Wu, Viktor Schlegel, Riza Batista-Navarro, Tharindu Madusanka, Iqra Zahid, Jiayan Zeng, Xiaochi Wang, Xinran He, Yizhi Li, and Goran Nenadic. 2024.
\newblock \href {https://doi.org/10.18653/v1/2024.findings-acl.9} {Which side are you on? a multi-task dataset for end-to-end argument summarisation and evaluation}.
\newblock In \emph{Findings of the Association for Computational Linguistics: ACL 2024}, pages 133--150, Bangkok, Thailand. Association for Computational Linguistics.

\bibitem[{Lin(2004)}]{lin-2004-rouge}
Chin-Yew Lin. 2004.
\newblock \href {https://aclanthology.org/W04-1013} {{ROUGE}: A package for automatic evaluation of summaries}.
\newblock In \emph{Text Summarization Branches Out}, pages 74--81, Barcelona, Spain. Association for Computational Linguistics.

\bibitem[{Liu et~al.(2023)Liu, Iter, Xu, Wang, Xu, and Zhu}]{liu-etal-2023-g}
Yang Liu, Dan Iter, Yichong Xu, Shuohang Wang, Ruochen Xu, and Chenguang Zhu. 2023.
\newblock \href {https://doi.org/10.18653/v1/2023.emnlp-main.153} {{G}-eval: {NLG} evaluation using gpt-4 with better human alignment}.
\newblock In \emph{Proceedings of the 2023 Conference on Empirical Methods in Natural Language Processing}, pages 2511--2522, Singapore. Association for Computational Linguistics.

\bibitem[{McInnes et~al.(2017)McInnes, Healy, and Astels}]{McInnes2017hdbscan}
Leland McInnes, John Healy, and Steve Astels. 2017.
\newblock \href {https://doi.org/10.21105/joss.00205} {hdbscan: Hierarchical density based clustering}.
\newblock \emph{Journal of Open Source Software}, 2(11):205.

\bibitem[{McInnes et~al.(2018)McInnes, Healy, Saul, and Großberger}]{McInnes2018umap}
Leland McInnes, John Healy, Nathaniel Saul, and Lukas Großberger. 2018.
\newblock \href {https://doi.org/10.21105/joss.00861} {Umap: Uniform manifold approximation and projection}.
\newblock \emph{Journal of Open Source Software}, 3(29):861.

\bibitem[{Misra et~al.(2016)Misra, Ecker, and Walker}]{misra-etal-2016-measuring}
Amita Misra, Brian Ecker, and Marilyn Walker. 2016.
\newblock \href {https://doi.org/10.18653/v1/W16-3636} {Measuring the similarity of sentential arguments in dialogue}.
\newblock In \emph{Proceedings of the 17th Annual Meeting of the Special Interest Group on Discourse and Dialogue}, pages 276--287, Los Angeles. Association for Computational Linguistics.

\bibitem[{Moratanch and Chitrakala(2017)}]{7944061}
N.~Moratanch and S.~Chitrakala. 2017.
\newblock \href {https://doi.org/10.1109/ICCCSP.2017.7944061} {A survey on extractive text summarization}.
\newblock In \emph{2017 International Conference on Computer, Communication and Signal Processing (ICCCSP)}, pages 1--6.

\bibitem[{Novikova et~al.(2017)Novikova, Du{\v{s}}ek, Cercas~Curry, and Rieser}]{novikova-etal-2017-need}
Jekaterina Novikova, Ond{\v{r}}ej Du{\v{s}}ek, Amanda Cercas~Curry, and Verena Rieser. 2017.
\newblock \href {https://doi.org/10.18653/v1/D17-1238} {Why we need new evaluation metrics for {NLG}}.
\newblock In \emph{Proceedings of the 2017 Conference on Empirical Methods in Natural Language Processing}, pages 2241--2252, Copenhagen, Denmark. Association for Computational Linguistics.

\bibitem[{Page et~al.(1998)Page, Brin, Motwani, and Winograd}]{Page1998PageRank}
Lawrence Page, Sergey Brin, Rajeev Motwani, and Terry Winograd. 1998.
\newblock \href {http://citeseerx.ist.psu.edu/viewdoc/summary?doi=10.1.1.31.1768} {{The PageRank Citation Ranking: Bringing Order to the Web}}.
\newblock Technical report, Stanford Digital Library Technologies Project.

\bibitem[{Papineni et~al.(2002)Papineni, Roukos, Ward, and Zhu}]{10.3115/1073083.1073135}
Kishore Papineni, Salim Roukos, Todd Ward, and Wei-Jing Zhu. 2002.
\newblock \href {https://doi.org/10.3115/1073083.1073135} {Bleu: a method for automatic evaluation of machine translation}.
\newblock In \emph{Proceedings of the 40th Annual Meeting on Association for Computational Linguistics}, ACL '02, page 311–318, USA. Association for Computational Linguistics.

\bibitem[{Peeperkorn et~al.()Peeperkorn, Kouwenhoven, Brown, and Jordanous}]{peeperkorntemperature}
Max Peeperkorn, Tom Kouwenhoven, Dan Brown, and Anna Jordanous.
\newblock Is temperature the creativity parameter of large language models?

\bibitem[{Qwen et~al.(2025)Qwen, :, Yang, Yang, Zhang, Hui, Zheng, Yu, Li, Liu, Huang, Wei, Lin, Yang, Tu, Zhang, Yang, Yang, Zhou, Lin, Dang, Lu, Bao, Yang, Yu, Li, Xue, Zhang, Zhu, Men, Lin, Li, Tang, Xia, Ren, Ren, Fan, Su, Zhang, Wan, Liu, Cui, Zhang, and Qiu}]{qwen2025qwen25technicalreport}
Qwen, :, An~Yang, Baosong Yang, Beichen Zhang, Binyuan Hui, Bo~Zheng, Bowen Yu, Chengyuan Li, Dayiheng Liu, Fei Huang, Haoran Wei, Huan Lin, Jian Yang, Jianhong Tu, Jianwei Zhang, Jianxin Yang, Jiaxi Yang, Jingren Zhou, Junyang Lin, Kai Dang, Keming Lu, Keqin Bao, Kexin Yang, Le~Yu, Mei Li, Mingfeng Xue, Pei Zhang, Qin Zhu, Rui Men, Runji Lin, Tianhao Li, Tianyi Tang, Tingyu Xia, Xingzhang Ren, Xuancheng Ren, Yang Fan, Yang Su, Yichang Zhang, Yu~Wan, Yuqiong Liu, Zeyu Cui, Zhenru Zhang, and Zihan Qiu. 2025.
\newblock \href {http://arxiv.org/abs/2412.15115} {Qwen2.5 technical report}.

\bibitem[{Radev et~al.(2002)Radev, Hovy, and McKeown}]{10.1162/089120102762671927}
Dragomir~R. Radev, Eduard Hovy, and Kathleen McKeown. 2002.
\newblock \href {https://doi.org/10.1162/089120102762671927} {Introduction to the special issue on summarization}.
\newblock \emph{Computational Linguistics}, 28(4):399--408.

\bibitem[{Reimers and Gurevych(2019)}]{reimers-gurevych-2019-sentence}
Nils Reimers and Iryna Gurevych. 2019.
\newblock \href {https://doi.org/10.18653/v1/D19-1410} {Sentence-{BERT}: Sentence embeddings using {S}iamese {BERT}-networks}.
\newblock In \emph{Proceedings of the 2019 Conference on Empirical Methods in Natural Language Processing and the 9th International Joint Conference on Natural Language Processing (EMNLP-IJCNLP)}, pages 3982--3992, Hong Kong, China. Association for Computational Linguistics.

\bibitem[{Reimers et~al.(2019)Reimers, Schiller, Beck, Daxenberger, Stab, and Gurevych}]{reimers-etal-2019-classification}
Nils Reimers, Benjamin Schiller, Tilman Beck, Johannes Daxenberger, Christian Stab, and Iryna Gurevych. 2019.
\newblock \href {https://doi.org/10.18653/v1/P19-1054} {Classification and clustering of arguments with contextualized word embeddings}.
\newblock In \emph{Proceedings of the 57th Annual Meeting of the Association for Computational Linguistics}, pages 567--578, Florence, Italy. Association for Computational Linguistics.

\bibitem[{Schiller et~al.(2021)Schiller, Daxenberger, and Gurevych}]{schiller-etal-2021-aspect}
Benjamin Schiller, Johannes Daxenberger, and Iryna Gurevych. 2021.
\newblock \href {https://doi.org/10.18653/v1/2021.naacl-main.34} {Aspect-controlled neural argument generation}.
\newblock In \emph{Proceedings of the 2021 Conference of the North American Chapter of the Association for Computational Linguistics: Human Language Technologies}, pages 380--396, Online. Association for Computational Linguistics.

\bibitem[{Sellam et~al.(2020)Sellam, Das, and Parikh}]{sellam-etal-2020-bleurt}
Thibault Sellam, Dipanjan Das, and Ankur Parikh. 2020.
\newblock \href {https://doi.org/10.18653/v1/2020.acl-main.704} {{BLEURT}: Learning robust metrics for text generation}.
\newblock In \emph{Proceedings of the 58th Annual Meeting of the Association for Computational Linguistics}, pages 7881--7892, Online. Association for Computational Linguistics.

\bibitem[{Sethi et~al.(2017)Sethi, Sonawane, Khanwalker, and Keskar}]{8336568}
Prakhar Sethi, Sameer Sonawane, Saumitra Khanwalker, and R.~B. Keskar. 2017.
\newblock \href {https://doi.org/10.1109/BID.2017.8336568} {Automatic text summarization of news articles}.
\newblock In \emph{2017 International Conference on Big Data, IoT and Data Science (BID)}, pages 23--29.

\bibitem[{Wang and Ling(2016)}]{wang-ling-2016-neural}
Lu~Wang and Wang Ling. 2016.
\newblock \href {https://doi.org/10.18653/v1/N16-1007} {Neural network-based abstract generation for opinions and arguments}.
\newblock In \emph{Proceedings of the 2016 Conference of the North {A}merican Chapter of the Association for Computational Linguistics: Human Language Technologies}, pages 47--57, San Diego, California. Association for Computational Linguistics.

\bibitem[{Wang et~al.(2024)Wang, Yu, Zeng, Yang, Wang, Chen, Jiang, Xie, Wang, Xie, Ye, Zhang, and Zhang}]{pandalm2024}
Yidong Wang, Zhuohao Yu, Zhengran Zeng, Linyi Yang, Cunxiang Wang, Hao Chen, Chaoya Jiang, Rui Xie, Jindong Wang, Xing Xie, Wei Ye, Shikun Zhang, and Yue Zhang. 2024.
\newblock Pandalm: An automatic evaluation benchmark for llm instruction tuning optimization.

\bibitem[{Xu et~al.(2023)Xu, Wang, Pan, Song, Freitag, Wang, and Li}]{xu-etal-2023-instructscore}
Wenda Xu, Danqing Wang, Liangming Pan, Zhenqiao Song, Markus Freitag, William Wang, and Lei Li. 2023.
\newblock \href {https://doi.org/10.18653/v1/2023.emnlp-main.365} {{INSTRUCTSCORE}: Towards explainable text generation evaluation with automatic feedback}.
\newblock In \emph{Proceedings of the 2023 Conference on Empirical Methods in Natural Language Processing}, pages 5967--5994, Singapore. Association for Computational Linguistics.

\bibitem[{Yang et~al.(2025)Yang, Li, Yang, Zhang, Hui, Zheng, Yu, Gao, Huang, Lv, Zheng, Liu, Zhou, Huang, Hu, Ge, Wei, Lin, Tang, Yang, Tu, Zhang, Yang, Yang, Zhou, Zhou, Lin, Dang, Bao, Yang, Yu, Deng, Li, Xue, Li, Zhang, Wang, Zhu, Men, Gao, Liu, Luo, Li, Tang, Yin, Ren, Wang, Zhang, Ren, Fan, Su, Zhang, Zhang, Wan, Liu, Wang, Cui, Zhang, Zhou, and Qiu}]{yang2025qwen3technicalreport}
An~Yang, Anfeng Li, Baosong Yang, Beichen Zhang, Binyuan Hui, Bo~Zheng, Bowen Yu, Chang Gao, Chengen Huang, Chenxu Lv, Chujie Zheng, Dayiheng Liu, Fan Zhou, Fei Huang, Feng Hu, Hao Ge, Haoran Wei, Huan Lin, Jialong Tang, Jian Yang, Jianhong Tu, Jianwei Zhang, Jianxin Yang, Jiaxi Yang, Jing Zhou, Jingren Zhou, Junyang Lin, Kai Dang, Keqin Bao, Kexin Yang, Le~Yu, Lianghao Deng, Mei Li, Mingfeng Xue, Mingze Li, Pei Zhang, Peng Wang, Qin Zhu, Rui Men, Ruize Gao, Shixuan Liu, Shuang Luo, Tianhao Li, Tianyi Tang, Wenbiao Yin, Xingzhang Ren, Xinyu Wang, Xinyu Zhang, Xuancheng Ren, Yang Fan, Yang Su, Yichang Zhang, Yinger Zhang, Yu~Wan, Yuqiong Liu, Zekun Wang, Zeyu Cui, Zhenru Zhang, Zhipeng Zhou, and Zihan Qiu. 2025.
\newblock \href {http://arxiv.org/abs/2505.09388} {Qwen3 technical report}.

\bibitem[{Yuan et~al.(2021)Yuan, Neubig, and Liu}]{bartscore}
Weizhe Yuan, Graham Neubig, and Pengfei Liu. 2021.
\newblock \href {https://proceedings.neurips.cc/paper_files/paper/2021/file/e4d2b6e6fdeca3e60e0f1a62fee3d9dd-Paper.pdf} {Bartscore: Evaluating generated text as text generation}.
\newblock In \emph{Advances in Neural Information Processing Systems}, volume~34, pages 27263--27277. Curran Associates, Inc.

\bibitem[{Zhang et~al.(2018)Zhang, Li, and Yao}]{ZHANG201888}
Junsheng Zhang, Kun Li, and Changqing Yao. 2018.
\newblock \href {https://doi.org/https://doi.org/10.1016/j.procs.2018.03.052} {Event-based summarization for scientific literature in chinese}.
\newblock \emph{Procedia Computer Science}, 129:88--92.
\newblock 2017 INTERNATIONAL CONFERENCE ON IDENTIFICATION,INFORMATION AND KNOWLEDGEIN THE INTERNET OF THINGS.

\bibitem[{Zhang et~al.(2020)Zhang, Kishore, Wu, Weinberger, and Artzi}]{bertscore}
Tianyi Zhang, Varsha Kishore, Felix Wu, Kilian~Q. Weinberger, and Yoav Artzi. 2020.
\newblock \href {https://openreview.net/forum?id=SkeHuCVFDr} {Bertscore: Evaluating text generation with bert}.
\newblock In \emph{International Conference on Learning Representations}.

\bibitem[{Zhao et~al.(2019)Zhao, Peyrard, Liu, Gao, Meyer, and Eger}]{zhao-etal-2019-moverscore}
Wei Zhao, Maxime Peyrard, Fei Liu, Yang Gao, Christian~M. Meyer, and Steffen Eger. 2019.
\newblock \href {https://doi.org/10.18653/v1/D19-1053} {{M}over{S}core: Text generation evaluating with contextualized embeddings and earth mover distance}.
\newblock In \emph{Proceedings of the 2019 Conference on Empirical Methods in Natural Language Processing and the 9th International Joint Conference on Natural Language Processing (EMNLP-IJCNLP)}, pages 563--578, Hong Kong, China. Association for Computational Linguistics.

\bibitem[{Zhu et~al.(2023)Zhu, Wang, and Wang}]{zhu2023judgelm}
Lianghui Zhu, Xinggang Wang, and Xinlong Wang. 2023.
\newblock \href {http://arxiv.org/abs/2310.17631} {Judgelm: Fine-tuned large language models are scalable judges}.

\bibitem[{Ziegenbein et~al.(2024)Ziegenbein, Syed, Potthast, and Wachsmuth}]{10.1007/978-3-031-63536-6_20}
Timon Ziegenbein, Shahbaz Syed, Martin Potthast, and Henning Wachsmuth. 2024.
\newblock Objective argument summarization in search.
\newblock In \emph{Robust Argumentation Machines}, pages 335--351, Cham. Springer Nature Switzerland.

\end{thebibliography}
